\pdfoutput=1

\documentclass[11pt]{article}

\usepackage[final]{acl}

\usepackage{times}
\usepackage{latexsym}

\usepackage[T1]{fontenc}

\usepackage[utf8]{inputenc}

\usepackage{microtype}

\usepackage{inconsolata}

\usepackage{graphicx}

\usepackage{multirow}
\usepackage{amssymb}
\usepackage{makecell}
\usepackage{amsmath}
\usepackage{relsize}
\usepackage{arydshln}
\usepackage{booktabs}
\usepackage{bm}
\usepackage{hyperref}

\usepackage{url}

\usepackage{amsfonts}
\usepackage{graphicx}
\usepackage{algorithm}
\usepackage{algorithmic}
\usepackage{tabu}
\usepackage{subfigure}

\usepackage{xcolor}
\usepackage{wrapfig} 
\usepackage{colortbl}
\usepackage[most]{tcolorbox}

\usepackage[normalem]{ulem}
\useunder{\uline}{\ul}{}

\definecolor{wkyellow}{RGB}{255,241,177}
\definecolor{lightgray}{HTML}{CCE5FF}
\definecolor{lightgray}{gray}{0.9}
\definecolor{goodblue}{HTML}{0071bc}

\makeatletter
\renewcommand{\maketag@@@}[1]{\hbox{\m@th\normalsize\normalfont#1}}%
\makeatother

\makeatletter
\def\adl@drawiv#1#2#3{%
        \hskip.5\tabcolsep
        \xleaders#3{#2.5\@tempdimb #1{1}#2.5\@tempdimb}%
                #2\z@ plus1fil minus1fil\relax
        \hskip.5\tabcolsep}
\newcommand{\cdashlinelr}[1]{%
  \noalign{\vskip\aboverulesep
           \global\let\@dashdrawstore\adl@draw
           \global\let\adl@draw\adl@drawiv}
  \cdashline{#1}
  \noalign{\global\let\adl@draw\@dashdrawstore
           \vskip\belowrulesep}}
\makeatother

\title{Improving MLLM's Document Image Machine Translation via Synchronously Self-reviewing Its OCR Proficiency}

\author{
\normalsize{Yupu Liang}\textsuperscript{1,2},
\normalsize{Yaping Zhang}\textsuperscript{1,2},
\normalsize{Zhiyang Zhang}\textsuperscript{1,2},
\normalsize{Zhiyuan Chen}\textsuperscript{1,2}, \\
{\bf \normalsize{Yang Zhao}\textsuperscript{1,2}},
{\bf \normalsize{Lu Xiang}\textsuperscript{1,2}},
{\bf \normalsize{Chengqing Zong}\textsuperscript{1,2}},
{\bf \normalsize{Yu Zhou}\textsuperscript{1,3}}\thanks{\ \ Corresponding author.}\ \  \\
\textsuperscript{1} \normalsize{State Key Laboratory of Multimodal Artificial Intelligence Systems (MAIS),} \\ \normalsize{Institute of Automation, Chinese Academy of Sciences, Beijing, China} \\
\textsuperscript{2} \normalsize{School of Artificial Intelligence, University of Chinese Academy of Sciences, Beijing, China} \\
\textsuperscript{3} \normalsize{Fanyu AI Laboratory, Zhongke Fanyu Technology Co., Ltd, Beijing, China} \\
\text{\small{\{liangyupu2021, zhangzhiyang2020, chenzhiyuan2023\}@ia.ac.cn, }}\small{\{yaping.zhang, yang.zhao, lu.xiang, cqzong, yzhou\}@nlpr.ia.ac.cn}
}

\begin{document}
\maketitle
\begin{abstract}

Multimodal Large Language Models (MLLMs) have shown strong performance in document image tasks, especially Optical Character Recognition (OCR). 
However, they struggle with Document Image Machine Translation (DIMT), which requires handling both cross-modal and cross-lingual challenges. 
Previous efforts to enhance DIMT capability through Supervised Fine-Tuning (SFT) on the DIMT dataset often result in the forgetting of the model's existing monolingual abilities, such as OCR.
To address these challenges, we introduce a novel fine-tuning paradigm, named \textbf{S}ynchronously \textbf{S}elf-\textbf{R}eviewing (\textbf{SSR}) its OCR proficiency, inspired by the concept "\textit{Bilingual Cognitive Advantage}". 
Specifically, SSR prompts the model to generate OCR text before producing translation text, which allows the model to leverage its strong monolingual OCR ability while learning to translate text across languages. 
Comprehensive experiments demonstrate the proposed SSR learning helps mitigate catastrophic forgetting, improving the generalization ability of MLLMs on both OCR and DIMT tasks.\footnote{Our code is available at: \url{https://github.com/liangyupu/SSR}}


\end{abstract}

\section{Introduction}

Multimodal Large Language Models (MLLMs) have achieved significant advancements in various document image understanding tasks, particularly in Optical Character Recognition (OCR), which plays a crucial role in extracting text from scanned documents or images. These improvements have led to notable progress in tasks, such as Visual Question Answering (VQA), and Information Extraction (IE) \citep{DBLP:journals/corr/abs-2401-12503, DBLP:journals/corr/abs-2403-04473, DBLP:journals/corr/abs-2409-12191, DBLP:conf/eccv/WeiKCZGYSHZ24}. However, MLLMs still face challenges towards Document Image Machine Translation (DIMT)—the task of translating text in document images from one language to another. \citep{zhang2023novel, zhang-etal-2023-layoutdit, liang-etal-2024-document}.

\begin{figure}[t]
    \centering
    \includegraphics[width=\columnwidth]{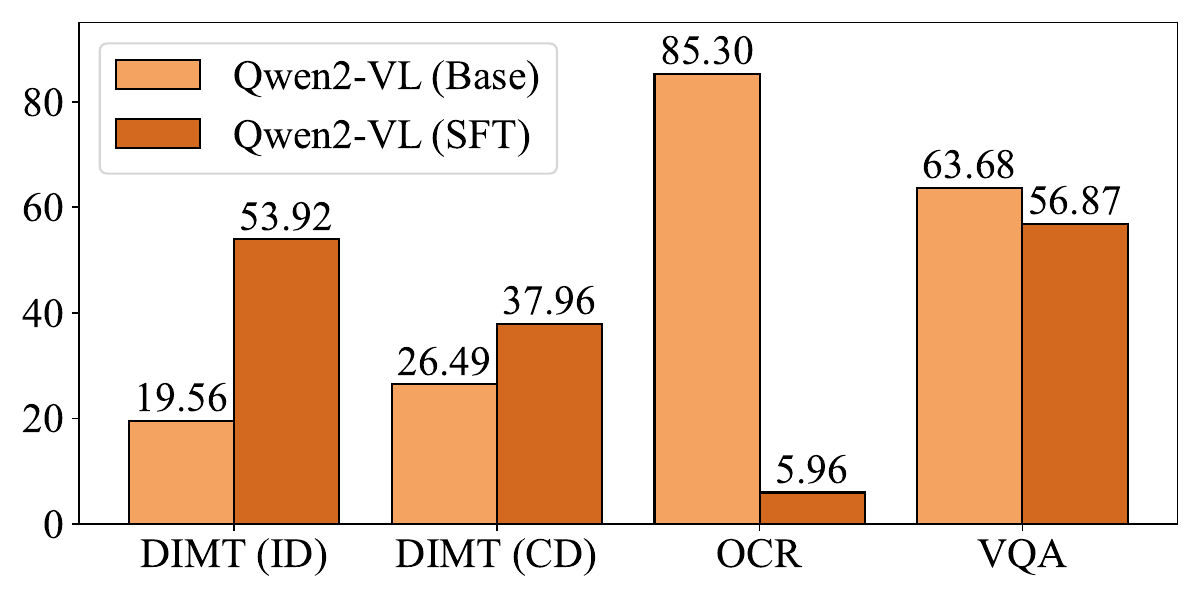}
    \caption{Performance of Qwen2-VL across various benchmarks. \textbf{Base} refers to the performance of the original MLLM, while \textbf{SFT} denotes the MLLM after fine-tuning on the DIMT dataset. \textbf{DIMT (ID)} and \textbf{DIMT (CD)} denote in-domain and cross-domain test separately. The evaluation metrics for DIMT, OCR, and VQA are BLEU, Character Accuracy (CA), and Average Normalized Levenshtein Similarity (ANLS), respectively.}
    \label{figure: introduction}
\end{figure}

An intuitive approach to enhancing MLLM's DIMT ability is to apply Supervised Fine-Tuning (SFT) \citep{DBLP:conf/nips/Ouyang0JAWMZASR22} on annotated DIMT datasets. 
However, a major challenge with SFT is catastrophic forgetting, where fine-tuning MLLM on translation tasks often causes a loss of the model's original OCR capability.
As shown in Figure~\ref{figure: introduction}, while the fine-tuned MLLM performs well on translation tasks, achieving a BLEU score of 53.92 on the in-domain DIMT task, it struggles to accurately extract text from images, with an accuracy of only 5.96 on the OCR task.
 This significant drop in OCR performance indicates a near-complete loss of the MLLM’s OCR proficiency.



To address the challenges associated with SFT, existing continual learning methods have been proposed \citep{yin-etal-2022-contintin, mok-etal-2023-large, yang-etal-2024-self, DBLP:journals/corr/abs-2404-16789, DBLP:journals/corr/abs-2410-13944}.
These methods aim to mitigate catastrophic forgetting and enhance domain generalization through various strategies, such as replay-based methods and regularization-based methods.
However, challenges persist in effectively balancing the retention of prior knowledge with the acquisition of new skills, especially in complex tasks like DIMT.

\begin{figure}[t]
    \centering
    \includegraphics[width=0.9\columnwidth]{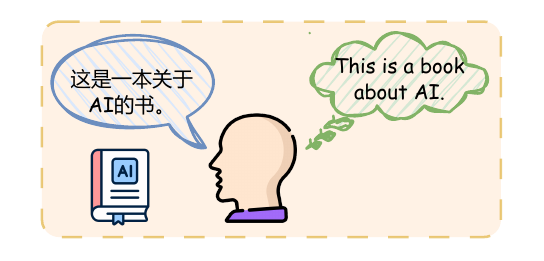}
    \caption{Bilingual individuals exhibit greater linguistic proficiency.}
    \label{figure: introduction pic}
\end{figure}

Inspired by the concept of "\textit{Bilingual Cognitive Advantage}" \citep{bialystok1991language, hamers1998cognitive, bialystok2001bilingualism, bialystok2010cognitive, zhang2023navigating, zhang2024mulcogbench, ZHANG2025121096}, as shown in Figure~\ref{figure: introduction pic}, a learning paradigm that focuses on retaining and leveraging human's existing monolingual strengths while learning new languages, we introduce a simple yet effective fine-tuning paradigm called \textbf{S}ynchronized \textbf{S}elf-\textbf{R}eviewing (\textbf{SSR}), where the MLLM generates the OCR text in the source language first, followed by the translation text in the target language. By synchronous learning, SSR enables the MLLM to leverage its strong monolingual OCR proficiency while extending its capabilities to new languages, thereby improving its cross-lingual performance on the DIMT task. Additionally, SSR enhances the MLLM's generalization ability, making it more robust across various domains and tasks. Furthermore, the method benefits from the use of large amounts of unsupervised data, reducing the need for extensive parallel datasets, which are often scarce in the DIMT task.

In summary, this paper presents a novel method to improve DIMT performance by using synchronously self-reviewing to preserve monolingual OCR proficiency while enabling cross-lingual DIMT. We demonstrate, through extensive experiments, that SSR significantly enhances the MLLM's generalization across both OCR and DIMT tasks, addressing challenges such as catastrophic forgetting and poor domain generalization.


Our contributions are summarized as follows:
\begin{itemize}
\item We propose a novel fine-tuning paradigm, SSR, which leverages the strong monolingual capabilities of MLLMs to enhance their cross-lingual performance.  
\item We introduce synchronous self-reviewing to utilize the MLLM's OCR proficiency and preserve its monolingual capability.  
\item Extensive experiments validate the effectiveness of the proposed method in improving the generalization ability of MLLMs on the DIMT task while maintaining their monolingual competence.
\end{itemize}

\section{Related Work}
Different from text machine translation \citep{yang2023bigtranslate, yang2024language, yang2025implicit}, document image machine translation aims to translate text within document images from one language to another while preserving the logical layout \citep{liang-etal-2024-document}.
Recent advancements in DIMT can be categorized into two primary approaches: (1) \textbf{Cascade systems} \citep{DBLP:conf/aaai/HinamiIYM21, sable2023doc, zhang2023novel, DBLP:journals/corr/abs-2310-12430}, which employ multiple models sequentially and encounter issues such as structural redundancy, error propagation, and high latency.
(2) \textbf{End-to-end models} \citep{DBLP:conf/icpr/MaZTHWZ022, zhu-etal-2023-peit, zhang-etal-2023-layoutdit, liang-etal-2024-document, ma-etal-2024-born, zhang-etal-2025-chaotic, DBLP:journals/pami/ZhangZLMXZZZ25, guan-etal-2025-trifine}, which streamline the process by optimizing a unified training objective, thereby improving structural efficiency.
These end-to-end methods are increasingly attracting researchers' attention.
\citet{zhu-etal-2023-peit} introduces an end-to-end TIMT framework that bridges the modality gap with pre-trained models.
\citet{liang-etal-2024-document} assembles multiple pre-trained models to complete the end-to-end DIMT task.
\citet{DBLP:journals/pami/ZhangZLMXZZZ25} proposes a framework to unify the geometric layout and logical layout of document images.
While these end-to-end methods have demonstrated satisfactory performance, their effectiveness is restricted to respective training domains, with limited cross-domain generalization.

\begin{figure*}[t]
    \centering
    \includegraphics[width=2\columnwidth]{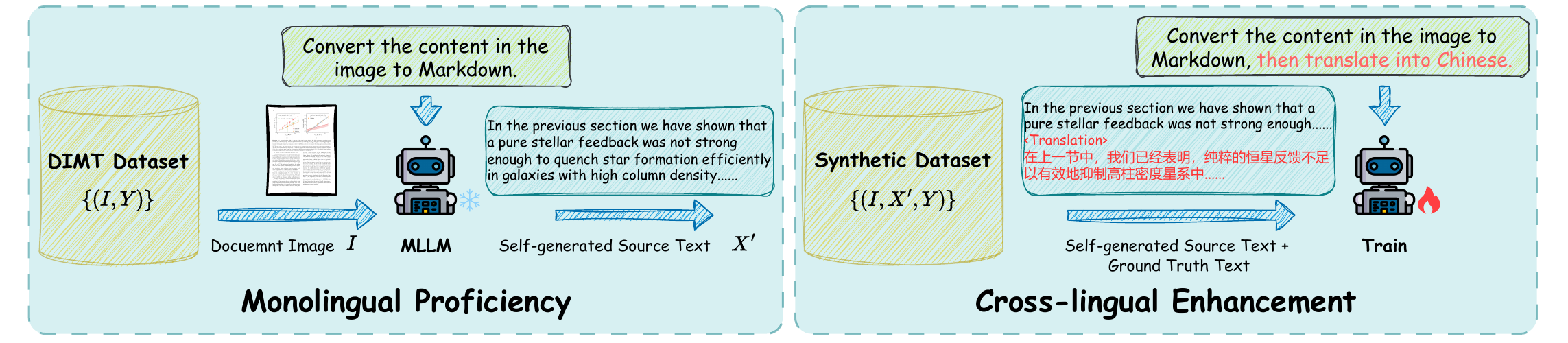}
    \caption{Overview of our proposed fine-tuning paradigm SSR. It contains two steps: (1) \textbf{Monolingual proficiency}: Given a document image and the original OCR prompt, the MLLM generates the source text (OCR result). (2) \textbf{Cross-lingual enhancement}: Use the self-generated source text and the ground truth target text to fine-tune the MLLM, enabling it to learn the relationship between the image, source text, and target text, while also smoothing the training process.}
    \label{figure: model}
\end{figure*}

Recent advancements in MLLMs have significantly improved the processing and understanding of text-rich document images \citep{hu-etal-2024-mplug, DBLP:journals/corr/abs-2409-03420, DBLP:journals/corr/abs-2401-12503, DBLP:conf/eccv/WeiKCZGYSHZ24, DBLP:journals/corr/abs-2403-04473, DBLP:journals/corr/abs-2404-09204, DBLP:journals/corr/abs-2409-12191, jian2024large, ren2025towards}.
\citet{DBLP:conf/eccv/WeiKCZGYSHZ24} explores adding fine-grained vision perception for document images to the MLLM without affecting its existing natural image understanding capabilities.
\citet{DBLP:journals/corr/abs-2403-04473} proposes shifted window attention to achieve cross-window connectivity at higher input resolutions and token resampler to filter out significant tokens.
\citet{DBLP:journals/corr/abs-2409-12191} introduces dynamic resolution mechanism and multimodal rotary position embeddings to facilitating the effective fusion of text, images, and videos.
Although MLLMs have demonstrated strong performance across various document image understanding tasks, their effectiveness diminishes for emerging tasks like DIMT.

\section{Method}
In this section, we will introduce SSR, a novel fine-tuning paradigm that leverage the MLLM's monolingual (OCR) proficiency to enhance its cross-lingual (DIMT) ability.
The overview of our approach is shown in Figure~\ref{figure: model}.
The key idea is to train the model to first generate the source text (OCR result) before producing the target text (translation text).
This approach enables the model to incorporate both image and source text information when generating the target text.
Although the self-generated source text may contain misrecognized or repeated text, since it is sampled from the model's original distribution, it contributes to a smoother convergence of the model's loss curve during training, which will be discussed in Section~\ref{section: main results}.
Furthermore, this self-review process helps in the retention of the model's original monolingual capabilities.

\subsection{Monolingual Proficiency}
This process involves prompting the MLLM with its original OCR instruction to perform OCR on the document image.
Since the generated text is sampled from the MLLM’s original distribution, it is better suited for maintaining its inherent monolingual capabilities.

Given a DIMT dataset $\mathcal{D} = \{(\bm{I}, \bm{Y})\}$, where $\bm{I}$ and $\bm{Y}$ denote the document image and corresponding ground truth target text, we prompt the MLLM to generate the OCR result $\bm{X}'$ for each document image $\bm{I}$ based on its original OCR instruction.
\begin{equation}
\small
    \bm{X}' \sim \mathrm{MLLM}(\bm{P}_{\mathrm{ocr}}, \bm{I})
\end{equation}

where $\bm{P}_{\mathrm{ocr}}$ is the MLLM's original OCR instruction.

This process is similar to some replay methods \citep{DBLP:journals/corr/abs-2404-16789} in continual learning; however, the key difference is that we allow the MLLM to generate its own replay data.

\subsection{Cross-lingual Enhancement}
This process concatenates the self-generated source text and ground truth target text to fine-tune the MLLM.
This approach enables the model to learn the relationship between different modalities while leveraging its monolingual capabilities to enhance cross-lingual performance, simultaneously facilitating self-review of its monolingual proficiency.

SSR constructs a prompt template based on the original OCR template.
Take Qwen2-VL \citep{DBLP:journals/corr/abs-2409-12191} as an example, the prompt construction is as follows:

\begin{tcolorbox}[colback=lightgray!50!white,colframe=lightgray,title=\textcolor{black}{SSR-constrained Prompt Template}, width=\columnwidth, breakable]
\footnotesize
\textbf{Instruction}: \newline
Convert the content in the image to Markdown \textcolor{gray}{(original OCR instruction of the MLLM)}, then translate into Chinese. \newline
\newline
\textbf{Response}: \newline
$\bm{X}'$ \textcolor{gray}{(self-generated source text)}\newline
<Translation> \textcolor{gray}{(special token)}\newline
$\bm{Y}$ \textcolor{gray}{(ground truth target text)}
\end{tcolorbox}

The constructed instruction-response pair is subsequently used to train the MLLM using the standard negative log-likelihood loss, which can be formulated as follows:
\begin{equation}
\small
    \mathcal{L} = -\sum_{t=1}^{r} \log p(\bm{R}_t | \bm{R}_{<t}, \bm{P}, \bm{I}; \bm{\theta})
\end{equation}
\begin{equation}
\small
    \bm{R} = \mathrm{CONCAT}(\bm{X}', \mathrm{<Translation>}, \bm{Y})
\end{equation}

where $\bm{R}_t$ denotes the $t$-th token of the response, $\bm{P}$ represents the instruction, $\bm{\theta}$ refers to the trainable parameters, and $r$ denotes the length of $\bm{R}$.

This approach trains the MLLM to gradually learn to generate target text, using the generated source text as a reference to guide target text generation.
This aligns more closely with the MLLM's original output distribution, resulting in a smoother training curve for the MLLM.

\section{Experiment}
\subsection{Dataset \& Metrics}
We randomly select 10K samples from the DoTA dataset \citep{liang-etal-2024-document} and comprehensively evaluate the model on the DoTA dataset for in-domain test and DITrans dataset \citep{zhang-etal-2023-layoutdit} for cross-domain test.
Detailed settings can be seen in Appendix~\ref{appendix: dataset settings}.

We thoroughly evaluate the models’ capabilities in three aspects: (1) \textbf{Full-text translation}, which means the translation quality of all the text in the image - BLEU.
(2) \textbf{Plain-text translation}, which means the translation quality of the text after removing formulas and tables - BLEU-PT.
(3) \textbf{Structure preserving}, which means the model’s ability to restore the layout structure of the document images - STEDS (Structure Tree-Edit-Distance-based Similarity).
All metric calculations follow the same procedure as described by \citet{liang-etal-2024-document}.

\subsection{Settings}
We select four MLLMs with different numbers of parameters: Vary-toy \citep{DBLP:journals/corr/abs-2401-12503}, Vary-base \citep{DBLP:conf/eccv/WeiKCZGYSHZ24}, Textmonkey \citep{DBLP:journals/corr/abs-2403-04473} and Qwen2-VL \citep{DBLP:journals/corr/abs-2409-12191}.
Given the constraints of our computational resources, the Low-Rank Adaptation (LoRA) technique \citep{DBLP:conf/iclr/HuSWALWWC22} is utilized in our experiments.
Specifically, a LoRA adapter with a rank of 16 is integrated into all the linear layers of the LLM part in the MLLM and exclusively trains the adapter.
The MLLMs are fine-tuned for 3 epochs on the train set.
We use the Adam optimizer and employ a linear decay learning rate schedule with a learning rate of 1e-4.
The batch is set to 32 for stable training.
The greedy search is used for inference.
More detailed settings are in Appendix~\ref{appendix: main experiment settings}.

\begin{table*}[t]
\renewcommand\arraystretch{0.97}
\footnotesize
\centering
\resizebox{\linewidth}{!}{
\begin{tabu}{lcccccccccc}
\toprule
\multicolumn{1}{c|}{\multirow{2}{*}{}} & \multicolumn{3}{c|}{\textbf{Academic Article (ID)}}                    & \multicolumn{3}{c|}{\textbf{Political Report (CD)}}                     & \multicolumn{3}{c|}{\textbf{Ads \& News (CD)}}                          & \textbf{Time}     \\
\multicolumn{1}{c|}{}                  & \textbf{BLEU}  & \textbf{BLEU-PT} & \multicolumn{1}{c|}{\textbf{STEDS}} & \textbf{BLEU}  & \textbf{BLEU-PT} & \multicolumn{1}{c|}{\textbf{STEDS}} & \textbf{BLEU}  & \textbf{BLEU-PT} & \multicolumn{1}{c|}{\textbf{STEDS}} & \textbf{s/page ($\downarrow$)} \\ \midrule
\multicolumn{11}{c}{\textbf{Baselines}}                                                                                                                                                                                                                                                     \\ \midrule
\multicolumn{1}{l|}{LARDIT}               & 35.58           & 41.75             & \multicolumn{1}{c|}{75.83}          & 14.66           & 16.58             & \multicolumn{1}{c|}{57.77}          & 1.64           & 1.71             & \multicolumn{1}{c|}{41.63}          & 12.46                  \\
\multicolumn{1}{l|}{Nougat-trans}               & 43.37           & 50.79             & \multicolumn{1}{c|}{88.16}          & 18.39           & 19.21             & \multicolumn{1}{c|}{52.12}          & 2.71           & 2.83             & \multicolumn{1}{c|}{40.53}          & 17.03                  \\
\multicolumn{1}{l|}{DIMTDA}               & 38.68           & 42.34             & \multicolumn{1}{c|}{84.44}          & 12.64           & 15.03             & \multicolumn{1}{c|}{60.86}          & 2.06           & 2.17             & \multicolumn{1}{c|}{40.75}          & 9.82                  \\
\multicolumn{1}{l|}{UMTIT}               & 37.40           & 40.02             & \multicolumn{1}{c|}{82.37}          & 10.06           & 10.67             & \multicolumn{1}{c|}{51.90}          & 2.77           & 2.08             & \multicolumn{1}{c|}{40.87}          & 14.76                  \\
\multicolumn{1}{l|}{MTKD}               & 37.32           & 39.96             & \multicolumn{1}{c|}{82.28}          & 13.24           & 15.33             & \multicolumn{1}{c|}{59.58}          & 2.42           & 2.39             & \multicolumn{1}{c|}{40.89}          & 9.24                  \\
\multicolumn{1}{l|}{AnyTrans}               & 32.98           & 34.94             & \multicolumn{1}{c|}{75.83}          & 31.05           & 31.05             & \multicolumn{1}{c|}{57.77}          & 16.47           & 17.89             & \multicolumn{1}{c|}{41.63}          & 14.81                  \\ \midrule

\multicolumn{11}{c}{\textbf{Vary-base (8.1B)}}                                                                                                                                                                                                                                                    \\ \midrule
\multicolumn{1}{l|}{Base}              & 13.45          & 5.79             & \multicolumn{1}{c|}{76.26}          & 2.84           & 2.79             & \multicolumn{1}{c|}{56.21}          & 1.06           & 1.06             & \multicolumn{1}{c|}{44.17}          & \textbf{47.62}                  \\ \cdashlinelr{1-11}
\multicolumn{1}{l|}{CoT (Direct)}               & 11.41           & 4.60             & \multicolumn{1}{c|}{79.89}          & 2.37           & 2.31             & \multicolumn{1}{c|}{57.11}          & 0.95           & 0.96             & \multicolumn{1}{c|}{\textbf{51.05}}          & 52.32                  \\
\multicolumn{1}{l|}{CoT (Cascade)}               & 3.42           & 1.81             & \multicolumn{1}{c|}{42.11}          & 2.90           & 2.73             & \multicolumn{1}{c|}{41.17}          & 0.87           & 0.87             & \multicolumn{1}{c|}{37.14}          & 120.54                  \\
\multicolumn{1}{l|}{SFT (MT)}          & 3.94           & 2.48             & \multicolumn{1}{c|}{48.00}          & 3.29           & 3.16             & \multicolumn{1}{c|}{\textbf{57.89}} & 1.18           & 1.18             & \multicolumn{1}{c|}{49.91} & 233.08                  \\
\multicolumn{1}{l|}{SFT (DIMT)}        & 19.84          & 18.60            & \multicolumn{1}{c|}{75.71}          & 4.46           & 4.49             & \multicolumn{1}{c|}{46.9}           & 0.94           & 0.94             & \multicolumn{1}{c|}{36.70}          & 92.25                  \\
\multicolumn{1}{l|}{SDFT}              & 11.56          & 11.51            & \multicolumn{1}{c|}{67.30}          & 2.99           & 3.02             & \multicolumn{1}{c|}{42.13}          & 0.79           & 0.82             & \multicolumn{1}{c|}{33.96}          & 137.93                  \\
\multicolumn{1}{l|}{SSR}       & \textbf{33.86} & \textbf{34.50}   & \multicolumn{1}{c|}{\textbf{81.72}}          & \textbf{21.47}          & \textbf{22.03}            & \multicolumn{1}{c|}{50.92}          & \textbf{6.68}  & \textbf{6.69}    & \multicolumn{1}{c|}{49.07}          & 150.44                  \\ \midrule
\multicolumn{11}{c}{\textbf{Textmonkey (9.7B)}}                                                                                                                                                                                                                                                   \\ \midrule
\multicolumn{1}{l|}{Base}              & 0.12           & 0.21             & \multicolumn{1}{c|}{29.37}          & 0.36           & 0.62             & \multicolumn{1}{c|}{31.90}          & 0.32           & 0.67             & \multicolumn{1}{c|}{26.65}          & \textbf{64.98}                  \\ \cdashlinelr{1-11}
\multicolumn{1}{l|}{CoT (Direct)}               & 0.34           & 0.33             & \multicolumn{1}{c|}{33.65}          & 0.99           & 0.94             & \multicolumn{1}{c|}{37.85}          & 0.88           & 0.48             & \multicolumn{1}{c|}{33.75}          & 71.88                  \\
\multicolumn{1}{l|}{CoT (Cascade)}               & 0.47           & 0.61             & \multicolumn{1}{c|}{29.43}          & 0.52           & 0.74             & \multicolumn{1}{c|}{31.90}          & 0.34           & 0.70             & \multicolumn{1}{c|}{26.69}          & 123.21                  \\
\multicolumn{1}{l|}{SFT (MT)}          & 16.69          & 18.93            & \multicolumn{1}{c|}{69.42}          & 12.26          & 12.26            & \multicolumn{1}{c|}{\textbf{61.06}}          & 5.26           & 5.26             & \multicolumn{1}{c|}{52.21}          & 259.22                  \\
\multicolumn{1}{l|}{SFT (DIMT)}        & 21.10          & 24.50            & \multicolumn{1}{c|}{73.07}          & 15.98          & 16.07            & \multicolumn{1}{c|}{60.46}          & 6.07           & 6.07             & \multicolumn{1}{c|}{54.25}          & 97.99                  \\
\multicolumn{1}{l|}{SDFT}              & 20.50          & 24.04            & \multicolumn{1}{c|}{71.80}           & 26.62          & 27.31            & \multicolumn{1}{c|}{58.51}          & 9.26           & 9.28             & \multicolumn{1}{c|}{\textbf{55.68}}          &  137.74                 \\
\multicolumn{1}{l|}{SSR}       & \textbf{26.45} & \textbf{28.55}   & \multicolumn{1}{c|}{\textbf{75.97}}          & \textbf{32.66} & \textbf{33.57}   & \multicolumn{1}{c|}{59.37}          & \textbf{12.40} & \textbf{12.40}   & \multicolumn{1}{c|}{54.31}          & 147.83                  \\ \midrule
\multicolumn{11}{c}{\textbf{Qwen2-VL (8.3B)}}                                                                                                                                                                                                                                                     \\ \midrule
\multicolumn{1}{l|}{Base}              & 19.56          & 15.38            & \multicolumn{1}{c|}{57.29}          & 26.49          & 26.51            & \multicolumn{1}{c|}{58.10}           & 11.19          & 11.19            & \multicolumn{1}{c|}{58.81}          & \textbf{33.58}                  \\ \cdashlinelr{1-11}
\multicolumn{1}{l|}{CoT (Direct)}               & 12.71          & 8.01            & \multicolumn{1}{c|}{57.94}          & 22.16          & 22.30            & \multicolumn{1}{c|}{61.34}           & 6.12          & 6.12            & \multicolumn{1}{c|}{57.89}          & 40.71                  \\
\multicolumn{1}{l|}{CoT (Cascade)}               & 29.44          & 27.07            & \multicolumn{1}{c|}{57.75}          & 36.37          & 36.31            & \multicolumn{1}{c|}{63.50}           & 28.92          & 28.92            & \multicolumn{1}{c|}{68.69}          & 58.69                  \\
\multicolumn{1}{l|}{SFT (MT)}          & 33.07          & 35.30            & \multicolumn{1}{c|}{63.91}          & 35.79          & 35.78            & \multicolumn{1}{c|}{64.17}          & 18.68          & 18.68            & \multicolumn{1}{c|}{50.67}          & 113.72                  \\
\multicolumn{1}{l|}{SFT (DIMT)}        & 53.92          & 53.20            & \multicolumn{1}{c|}{87.27}          & 37.96          & 37.93            & \multicolumn{1}{c|}{63.08}          & 23.48          & 23.49            & \multicolumn{1}{c|}{69.72}          & 51.64                  \\
\multicolumn{1}{l|}{SDFT}              & 53.55          & 55.11            & \multicolumn{1}{c|}{87.17}          & 39.01          & 38.97            & \multicolumn{1}{c|}{63.25}          & 27.65          & 27.65            & \multicolumn{1}{c|}{67.78}          & 54.26                  \\
\multicolumn{1}{l|}{SSR}       & \textbf{57.23} & \textbf{58.88}   & \multicolumn{1}{c|}{\textbf{89.65}} & \textbf{41.91} & \textbf{41.80}   & \multicolumn{1}{c|}{\textbf{67.28}} & \textbf{33.61} & \textbf{33.59}   & \multicolumn{1}{c|}{\textbf{71.98}} & 95.48                  \\ \bottomrule
\end{tabu}
}
\caption{Results on DoTA and DITrans dataset. All MLLMs are fine-tuned on the DoTA dataset and tested on both the DoTA dataset, which contains images from the \textbf{Academic Article} domain, serving as the \textbf{in-domain (ID)} test and the DITrans dataset, which includes images from the \textbf{Political Report}, \textbf{Ads \& News} domains, serving as the zero-shot \textbf{cross-domain (CD)} test. The number of parameters for each MLLM is provided alongside its respective model. The \textbf{Time} refers to the average inference time on a single NVIDIA A100 GPU. ($\downarrow$) indicates that for this metric, lower values are better. The \textbf{bold numbers} indicate the best performance achieved by each MLLM.}
\label{table: main experiment}
\end{table*}

\subsection{Baselines}
We evaluate our method against diverse baselines, including small models, MLLMs with Chain of Thought (CoT), Supervised Fine-tuning (SFT), and replay method, to comprehensively assess its performance and validate its effectiveness.

\noindent \textbf{$\bullet$ Small Model Baselines}

\textbf{LARDIT }\citep{zhang2023novel} This cascade system employs a layout analysis model \citep{DBLP:journals/corr/abs-2310-12430}, the \href{https://github.com/tesseract-ocr/tesseract}{OCR tool}, and a text-only machine translation model trained on the DoTA dataset, sequentially.

\textbf{Nougat-trans }\citep{DBLP:conf/iclr/BlecherCSS24} We utilize the Nougat model for combined layout analysis and OCR and the text-only machine translation model is employed for translation.

\textbf{DIMTDA }\citep{liang-etal-2024-document} This end-to-end DIMT model uses a model assembler to integrate multiple pre-trained models to enhance the understanding of layout and translation capabilities.

\textbf{UMTIT }\citep{DBLP:conf/coling/NiuM024} This model consists of two image-text modality conversion steps. We only use the result of the first step for evaluation, which converts images to text to recognize the source text and generate translations.

\textbf{MTKD }\citep{DBLP:conf/icdar/MaZTZZZ23} This method can effectively distillate knowledge from the pipeline model and utilizes three teacher models to improve the performance of the end-to-end TIMT model. 

\textbf{AnyTrans }\citep{qian-etal-2024-anytrans} This paper presents a framework entirely using open-source models, such as LLMs and text-guided diffusion models, to complete in-image machine translation. We only use the result of the translated text for evaluation.

The following lists the baselines based on MLLMs.
The detailed prompts for each method can be seen in Appendix~\ref{appendix: prompts}.

\textbf{Base } We directly prompt the original MLLM to perform the DIMT task.

\noindent \textbf{$\bullet$ CoT Baselines}

\textbf{CoT (Direct) }\citep{DBLP:conf/nips/Wei0SBIXCLZ22} We directly prompt the original MLLM to perform "OCR than translation" on the document image.

\textbf{CoT (Cascade) }\citep{DBLP:conf/nips/Wei0SBIXCLZ22} We first prompt the original MLLM to perform OCR, and then prompt it to generate the translation based on both the image and the OCR result.

\noindent \textbf{$\bullet$ SFT Baselines}

\textbf{SFT (MT) }\citep{DBLP:conf/nips/Ouyang0JAWMZASR22} The MLLM is first fine-tuned on the  English-Chinese parallel corpus from the training set, and then CoT (Cascade) method above is applied to generate translations.

\textbf{SFT (DIMT) }\citep{DBLP:conf/nips/Ouyang0JAWMZASR22} The MLLM is directly fine-tuned on the train set.

\noindent \textbf{$\bullet$ Replay Baseline}

\textbf{SDFT }\citep{yang-etal-2024-self} This method fine-tunes the model with a distilled dataset generated by the model itself to match its original distribution.

\section{Results \& Analysis}
\subsection{Main Results}
\label{section: main results}
Table~\ref{table: main experiment} reports the performance of all methods.
It can be observed that our method outperforms the baselines in terms of translation quality across all MLLMs with varying sizes and structures.
The results of Vary-toy experiment can be seen in the Appendix~\ref{appendix: vay-toy}.

\noindent \textbf{$\bullet$ MLLM with Limited Instruction-following Ability }
In the Vary-base and Textmonkey experiments, the performance of SSR significantly surpasses all other methods.
Take the Vary-base experiment as an example, the improvements are 14.02 BLEU in the in-domain test, and 17.01 BLEU and 5.74 BLEU in two zero-shot cross-domain tests, compared to SFT (DIMT).
These results show that our approach can be applied to larger MLLMs, thereby validating its effectiveness in enhancing translation quality and generalization.

\begin{table*}[t]
\footnotesize
\centering
\begin{tabu}{ll|cc|cc|ccc}
\toprule
\multicolumn{2}{c|}{\multirow{2}{*}{}}                                 & \multicolumn{2}{c|}{\textbf{OCR (Document)}} & \multicolumn{2}{c|}{\textbf{OCR (Scene)}} & \textbf{DocVQA} & \textbf{InfoVQA} & \textbf{ChartQA} \\
\multicolumn{2}{c|}{}                                                  & \textbf{CA}           & \textbf{WA}          & \textbf{CA}         & \textbf{WA}         & \textbf{ANLS}   & \textbf{ANLS}    & \textbf{ANLS}    \\ \midrule
\multicolumn{1}{l|}{\multirow{5}{*}{\textbf{Vary-toy}}}   & Base       & \textbf{68.46}        & \textbf{65.17}       & 45.74               & 41.73               & \textbf{47.76}  & {\ul 5.13}       & \textbf{7.87}    \\ \cdashlinelr{2-9}
\multicolumn{1}{l|}{}                                     & SFT (MT)   & 16.95                 & 13.10                & {\ul 49.09}         & {\ul 46.54}         & 27.97           & 1.16             & 2.07             \\
\multicolumn{1}{l|}{}                                     & SFT (DIMT) & 14.53                 & 8.81                 & 31.61               & 29.56               & 40.10           & 2.73             & 5.25             \\
\multicolumn{1}{l|}{}                                     & SDFT       & {\ul 64.80}           & {\ul 61.50}          & 43.02               & 39.67               & 42.59           & \textbf{5.45}    & {\ul 5.98}       \\
\multicolumn{1}{l|}{}                                     & SSR        & 61.10                 & 56.86                & \textbf{51.8}       & \textbf{46.66}      & {\ul 43.61}     & 4.02             & 4.92             \\ \midrule
\multicolumn{1}{l|}{\multirow{5}{*}{\textbf{Vary-base}}}  & Base       & \textbf{68.48}        & \textbf{64.91}       & \textbf{81.41}      & \textbf{76.83}      & \textbf{66.38}  & \textbf{12.75}   & \textbf{12.51}   \\ \cdashlinelr{2-9}
\multicolumn{1}{l|}{}                                     & SFT (MT)   & 50.39                 & 46.15                & 80.61               & 47.78               & 56.25           & 6.67             & 6.55             \\
\multicolumn{1}{l|}{}                                     & SFT (DIMT) & 47.65                 & 42.46                & 70.01               & 33.33               & 60.82           & 11.02            & 11.7             \\
\multicolumn{1}{l|}{}                                     & SDFT       & {\ul 67.35}           & {\ul 63.59}          & 54.91               & 49.41               & {\ul 62.68}     & {\ul 11.49}      & {\ul 12.46}      \\
\multicolumn{1}{l|}{}                                     & SSR        & 66.14                 & 62.45                & {\ul 81.26}         & {\ul 75.92}         & 62.19           & 10.15            & 9.78             \\ \midrule
\multicolumn{1}{l|}{\multirow{5}{*}{\textbf{Textmonkey}}} & Base       & \textbf{75.52}        & \textbf{70.56}       & \textbf{84.74}      & \textbf{79.40}       & \textbf{58.39}  & 22.21            & \textbf{8.69}    \\ \cdashlinelr{2-9}
\multicolumn{1}{l|}{}                                     & SFT (MT)   & 9.23                  & 7.49                 & 78.46               & 74.53               & 39.17           & 12.15            & 6.10             \\
\multicolumn{1}{l|}{}                                     & SFT (DIMT) & 8.98                  & 5.69                 & 72.86               & 68.82               & 52.52           & {\ul 22.58}      & 7.47             \\
\multicolumn{1}{l|}{}                                     & SDFT       & {\ul 73.51}           & {\ul 69.77}          & 57.96               & 52.67               & 39.78           & 20.45            & 7.38             \\
\multicolumn{1}{l|}{}                                     & SSR        & 72.76                 & 67.57                & {\ul 83.11}         & {\ul 78.59}         & {\ul 55.02}     & \textbf{22.78}   & {\ul 8.11}       \\ \midrule
\multicolumn{1}{l|}{\multirow{5}{*}{\textbf{Qwen2-VL}}}   & Base       & {\ul 85.30}           & {\ul 78.20}          & 70.29               & 64.75               & \textbf{93.55}  & \textbf{63.07}   & \textbf{63.68}   \\ \cdashlinelr{2-9}
\multicolumn{1}{l|}{}                                     & SFT (MT)   & 5.83                  & 3.08                 & 19.83               & 17.45               & 84.90           & 54.33            & 46.57            \\
\multicolumn{1}{l|}{}                                     & SFT (DIMT) & 5.96                  & 2.17                 & 33.47               & 31.06               & 88.98           & 57.57            & 56.87            \\
\multicolumn{1}{l|}{}                                     & SDFT       & \textbf{86.72}        & \textbf{80.12}       & {\ul 71.98}         & {\ul 67.84}         & 90.55           & 59.72            & 60.51            \\
\multicolumn{1}{l|}{}                                     & SSR        & 85.18                 & 78.12                & \textbf{82.03}      & \textbf{77.48}      & {\ul 92.47}     & {\ul 60.56}      & {\ul 61.37}      \\ \bottomrule
\end{tabu}
\caption{Results of MLLMs' monolingual ability preserving after fine-tuning with different methods. The \textbf{bold numbers} indicate the best performance achieved by each MLLM, and the \uline{underline numbers} are the second best.}
\label{table: monolingual ability}
\end{table*}

\noindent \textbf{$\bullet$ MLLM with Strong Instruction-following Ability }
In the Qwen2-VL experiment, the original MLLM achieves high translation quality (53.92 BLEU in the in-domain test) despite requiring only minimal training data, our proposed method still outperforms SFT (DIMT) by margins of 3.31 BLEU and 5.68 BLEU-PT.
Furthermore, in zero-shot cross-domain evaluations on ads \& news domains, SSR surpasses SFT (DIMT) by substantial increments of 10.13 BLEU and 10.10 BLEU-PT.
These findings demonstrate that our approach remains applicable to more advanced MLLMs exhibiting superior instruction-following capabilities, aligning with the ongoing research direction in MLLM development.
The output samples for the DIMT test can be seen in Appendix~\ref{appendix: output samples}.

\begin{figure}[t]
    \centering
    \includegraphics[width=\columnwidth]{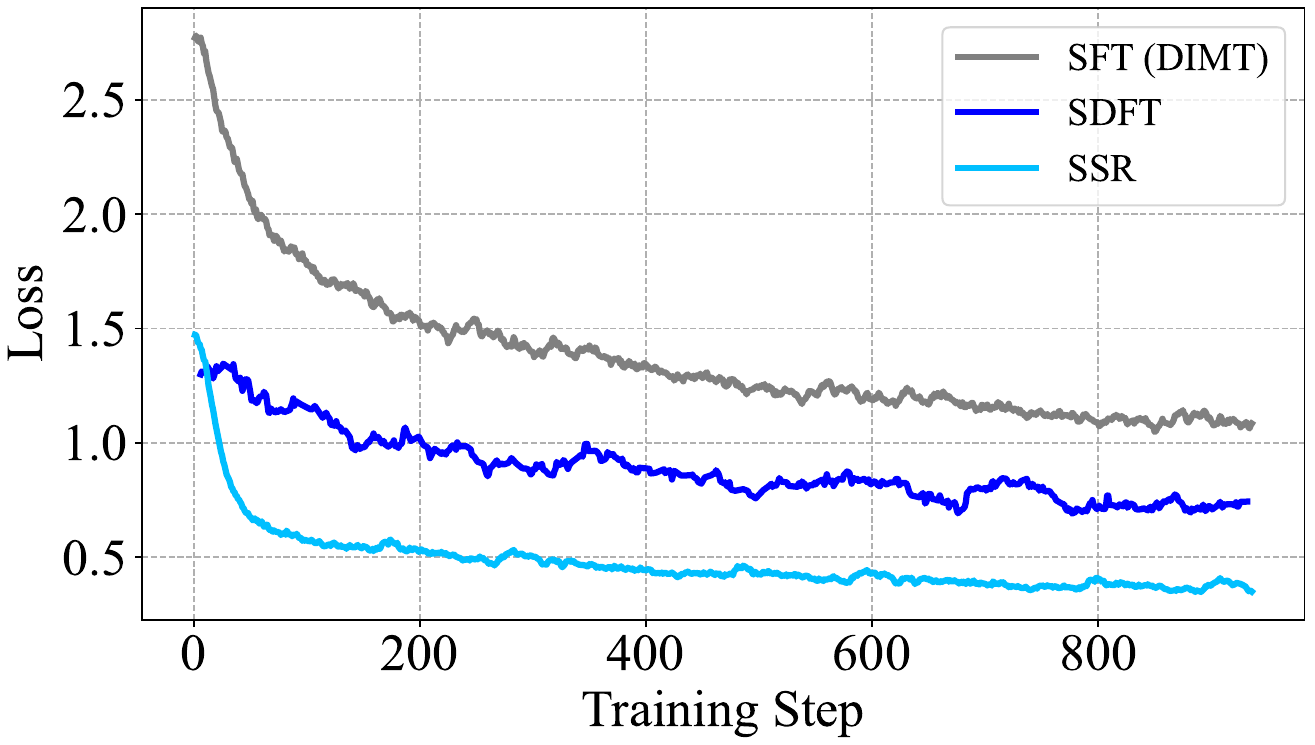}
    \caption{Training loss curves of different methods in the Vary-base experiment.}
    \label{figure: training loss}
     
\end{figure}

We also shows the training loss in the Vary-base experiment in Figure~\ref{figure: training loss}.
As shown in the figure, the training loss curve of SSR is smoother and achieves the lowest loss value.
This is due to the fact that the constructed data in the monolingual demonstration is sampled from the original distribution of Vary-base, making it more suitable for training.

\subsection{Monolingual Ability Preserving}

To assess the preservation of monolingual capabilities in base MLLMs across different methods, we perform a comprehensive evaluation using various benchmarks.
For OCR performance evaluation, we employ the DITrans dataset \citep{zhang2023novel} for document image testing and the FST dataset \citep{DBLP:conf/icdar/KaratzasGNGBIMN15} for scene text image testing, with Character Accuracy (CA) and Word Accuracy (WA) as quantitative metrics.
Visual Question Answering (VQA) capabilities are examined through the DocVQA \citep{DBLP:conf/wacv/MathewKJ21}, InfoVQA \citep{DBLP:conf/wacv/MathewBTKVJ22}, and ChartQA \citep{masry-etal-2022-chartqa} benchmarks, assessed via the Average Normalized Levenshtein Similarity (ANLS) metric. 
Notably, all evaluations are conducted in a zero-shot manner without additional fine-tuning on downstream task-specific datasets.
The results are listed in Table~\ref{table: monolingual ability}.

For OCR performance, both SFT-based methods result in a significant decline in OCR effectiveness across both scenarios, illustrating a classic case of catastrophic forgetting.
In contrast, SSR exhibits remarkable proficiency in maintaining the OCR capabilities of the base MLLMs.
Taking the Qwen2-VL experiment as an example, SSR causes only a 0.12 decrease in CA and a 0.08 decrease in WA in document image scenarios.
In the scene text image scenarios, SSR even surpasses the base MLLM, achieving a increase of 11.74 in CA and 12.73 in WA.
These results underscore the effectiveness of monolingual demonstrations in preserving the OCR capabilities of the base MLLMs.
In document image scenarios, SDFT achieves the best performance, as it is fine-tuned with document image OCR task data.
However, SSR still delivers comparable performance and surpasses SDFT in scene text image scenarios, highlighting its superior generalization capability.

In terms of VQA performance, our method also exhibits impressive preservation of monolingual abilities. 
In the Qwen2-VL experiment, the MLLM experiences only a 1.08 drop in ANLS on the DocVQA dataset, a negligible cost compared to the 4.57 ANLS drop seen with SFT (DIMT).
This highlights the effectiveness of our method in preserving unseen general monolingual capabilities.
The output samples for the OCR and VQA test can be seen in Appendix~\ref{appendix: output samples}.

\subsection{Cross-lingual Ability Generalization}

\begin{table}[t]
\footnotesize
\centering
\resizebox{\linewidth}{!}{
\begin{tabu}{ll|ccc}
\toprule
\multicolumn{2}{c|}{}                                            & \textbf{DocVQA} & \textbf{InfoVQA} & \textbf{ChartQA} \\ \midrule
\multicolumn{1}{l|}{\multirow{2}{*}{\textbf{Vary-toy}}}   & Base & 6.65            & 0.07             & 0.01             \\
\multicolumn{1}{l|}{}                                     & SSR  & 7.57            & 0.21             & 0.38             \\ \midrule
\multicolumn{1}{l|}{\multirow{2}{*}{\textbf{Vary-base}}}  & Base & 8.64            & 1.10             & 0.90             \\
\multicolumn{1}{l|}{}                                     & SSR  & 9.13            & 1.93             & 1.64             \\ \midrule
\multicolumn{1}{l|}{\multirow{2}{*}{\textbf{Textmonkey}}} & Base & 19.20           & 9.73             & 13.71            \\
\multicolumn{1}{l|}{}                                     & SSR  & 21.01           & 10.80            & 8.31             \\ \midrule
\multicolumn{1}{l|}{\multirow{2}{*}{\textbf{Qwen2-VL}}}   & Base & 46.99           & 38.32            & 50.27            \\
\multicolumn{1}{l|}{}                                     & SSR  & 55.16           & 40.32            & 50.37            \\ \bottomrule
\end{tabu}
}
\caption{Results of MLLMs' cross-lingual ability generalization after fine-tuning with SSR. The text in the input image is in English, while the questions and answers are in Chinese. The ANLS scores are reported.}
\label{table: cross-lingual ability}
\end{table}

In our preliminary experiments, we observed that MLLMs, after fine-tuning with SSR, generalize to cross-lingual document image understanding abilities.
Therefore, we conduct further comprehensive experiments to evaluate their cross-lingual capabilities.
We translate both the questions and answers in several VQA benchmarks into Chinese using \href{https://translate.google.com/}{Google Translate} and perform evaluation in a zero-shot manner.
The results are shown in Table~\ref{table: cross-lingual ability}.

It is evident that the cross-lingual document image understanding ability of MLLMs is significantly enhanced after fine-tuning with SSR.
Specifically, after fine-tuning, Qwen2-VL achieves improvements of 8.17 and 2.00 ANLS on the DocVQA and InfoVQA test sets, respectively.
Moreover, by comparing the performance of Vary-base, Textmonkey and Qwen2-VL, MLLMs with stronger instruction-following capabilities demonstrate more substantial improvements.
The output samples for the cross-lingual VQA test can be seen in Appendix~\ref{appendix: output samples}.

\begin{figure}[t]
    \centering
    \includegraphics[width=\columnwidth]{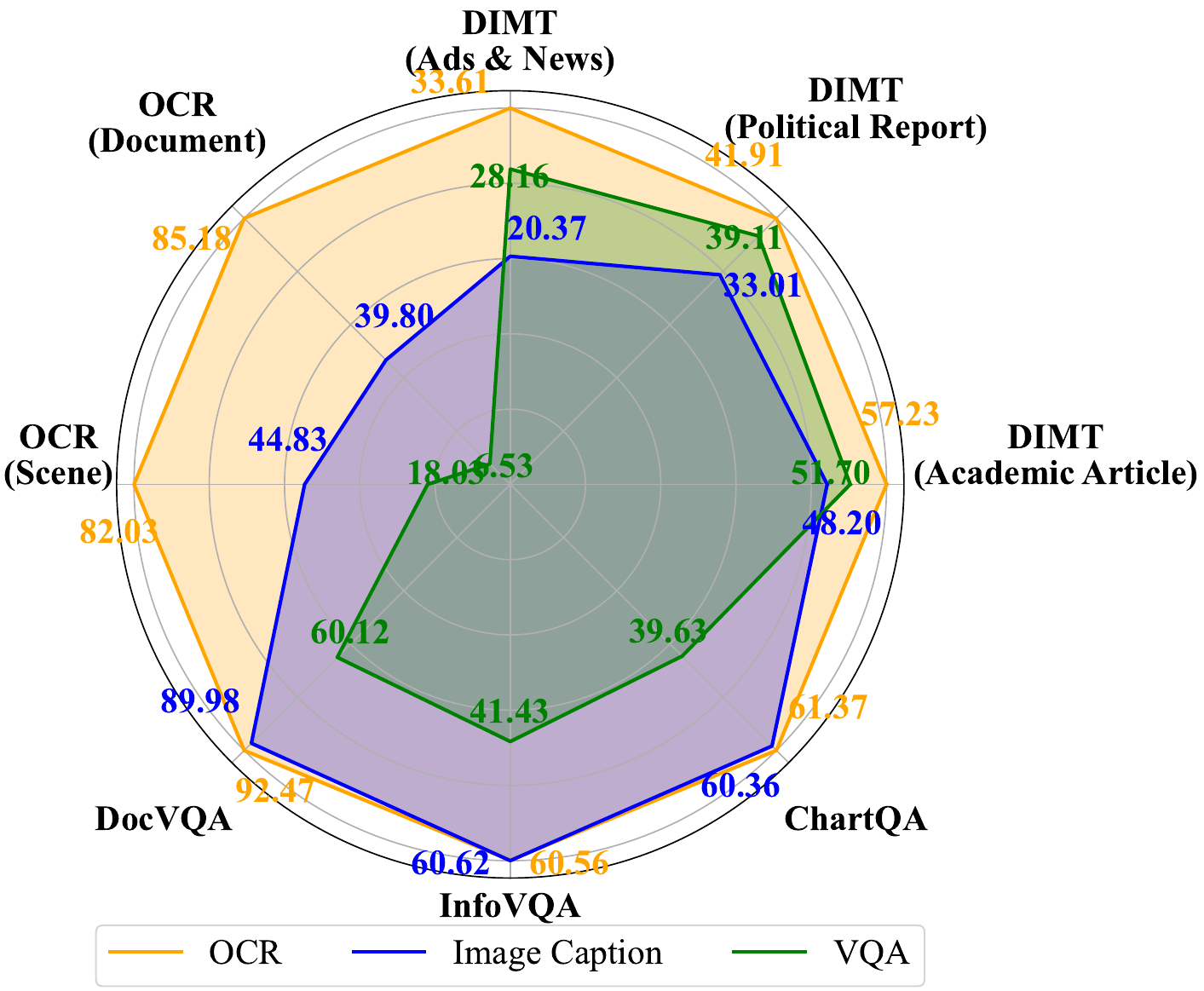}
    \caption{Results of Qwen2-VL through SSR fine-tuning using different monolingual tasks. Detailed data can be seen in Appendix~\ref{appendix: detailed data}. It is better to zoom in for a clearer view.}
    \label{figure: task selection}
     
\end{figure}

\begin{table*}[t]
\centering
\footnotesize
\resizebox{\linewidth}{!}{
\begin{tabular}{lccccccccc}
\toprule
\multicolumn{1}{c|}{\multirow{2}{*}{}} & \multicolumn{3}{c|}{\textbf{Academic Articles (ID)}}                          & \multicolumn{3}{c|}{\textbf{Political Report (CD)}}                    & \multicolumn{3}{c}{\textbf{Ads \& News (CD)}}     \\
\multicolumn{1}{c|}{}                  & \textbf{BLEU}        & \textbf{BLEU-PT} & \multicolumn{1}{c|}{\textbf{STEDS}} & \textbf{BLEU} & \textbf{BLEU-PT} & \multicolumn{1}{c|}{\textbf{STEDS}} & \textbf{BLEU} & \textbf{BLEU-PT} & \textbf{STEDS} \\ \midrule
\multicolumn{10}{c}{\textbf{Vary-base (8.1B)}}                                                                                                                                                                                                      \\ \midrule
\rowcolor{lightgray} \multicolumn{1}{l|}{Base}              & 13.45                & 5.79             & \multicolumn{1}{c|}{76.26}          & 2.84          & 2.79             & \multicolumn{1}{c|}{56.21}          & 1.06          & 1.06             & 44.17          \\ \cdashlinelr{1-10}
\multicolumn{1}{l|}{SSR w Ground Truth Text}          & 33.71                & 32.50            & \multicolumn{1}{c|}{\textbf{83.14}}          & \textbf{26.05}         & \textbf{26.90}            & \multicolumn{1}{c|}{\textbf{56.82}}          & 4.63          & 5.00             & 46.15          \\
\multicolumn{1}{l|}{SSR w OCR Text}          & 27.35                & 25.57            & \multicolumn{1}{c|}{72.16}          & 23.78         & 24.08            & \multicolumn{1}{c|}{52.50}          & 5.05          & 5.05             & 48.35          \\
\multicolumn{1}{l|}{SSR w Self-generated Text}          & \textbf{33.86}                & \textbf{34.50}            & \multicolumn{1}{c|}{81.72}          & 21.47         & 22.03            & \multicolumn{1}{c|}{50.92}          & \textbf{6.68}          & \textbf{6.69}             & \textbf{49.07}          \\ \midrule
\multicolumn{10}{c}{\textbf{Qwen2-VL (8.3B)}}                                                                                                                                                                                                       \\ \midrule
\rowcolor{lightgray} \multicolumn{1}{l|}{Base}              & 19.56                & 15.38            & \multicolumn{1}{c|}{57.29}          & 26.49         & 26.51            & \multicolumn{1}{c|}{58.1}           & 11.19         & 11.19            & 58.81          \\ \cdashlinelr{1-10}
\multicolumn{1}{l|}{SSR w Ground Truth Text}          & 54.55                & 58.07            & \multicolumn{1}{c|}{87.62}          & 41.43         & 41.38            & \multicolumn{1}{c|}{60.43}          & 32.55         & 32.55            & 68.14          \\
\multicolumn{1}{l|}{SSR w OCR Text}          & 52.83                & 52.03            & \multicolumn{1}{c|}{84.68}          & 37.63         & 38.24            & \multicolumn{1}{c|}{61.70}          & 29.09         & 29.09            & 64.08          \\
\multicolumn{1}{l|}{SSR w Self-generated Text}          & \textbf{57.23}                & \textbf{58.88}            & \multicolumn{1}{c|}{\textbf{89.65}}          & \textbf{41.91}         & \textbf{41.80}            & \multicolumn{1}{c|}{\textbf{67.28}}          & \textbf{33.61}         & \textbf{33.59}            & \textbf{71.98}          \\ \bottomrule
\end{tabular}
}
\caption{Results of Vary-base and Qwen2-VL through SSR fine-tuning using heterogeneous source texts. \textbf{ID} and \textbf{CD} denote in-domain and cross-domain test, respectively. The \textbf{bold numbers} indicate the best performance.}
\label{table: extension to GT OCR}
\end{table*}

\subsection{Monolingual Task Selection}
To investigate the impact of different monolingual tasks on our method, we select OCR, image caption, and VQA as the monolingual abilities to demonstrate, constructing synthetic data separately for fine-tuning Qwen2-VL.
Detailed prompt templates can be found in Appendix~\ref{appendix: prompts}.
All other settings remain consistent with the main experiment.
We use BLEU, CA, and ANLS as metrics.

Figure~\ref{figure: task selection} shows that using OCR as the demonstration task yields the best performance across all test sets, effectively enhancing cross-lingual ability while preserving monolingual proficiency.
We believe this is because, to complete the OCR task, the MLLM needs to generate the longest text, thereby preserving the most information from the original MLLM's output distribution while also providing more context for generating target text.

\subsection{Extension to Unsupervised Data}

\begin{figure}[t]
    \centering
    \includegraphics[width=\columnwidth]{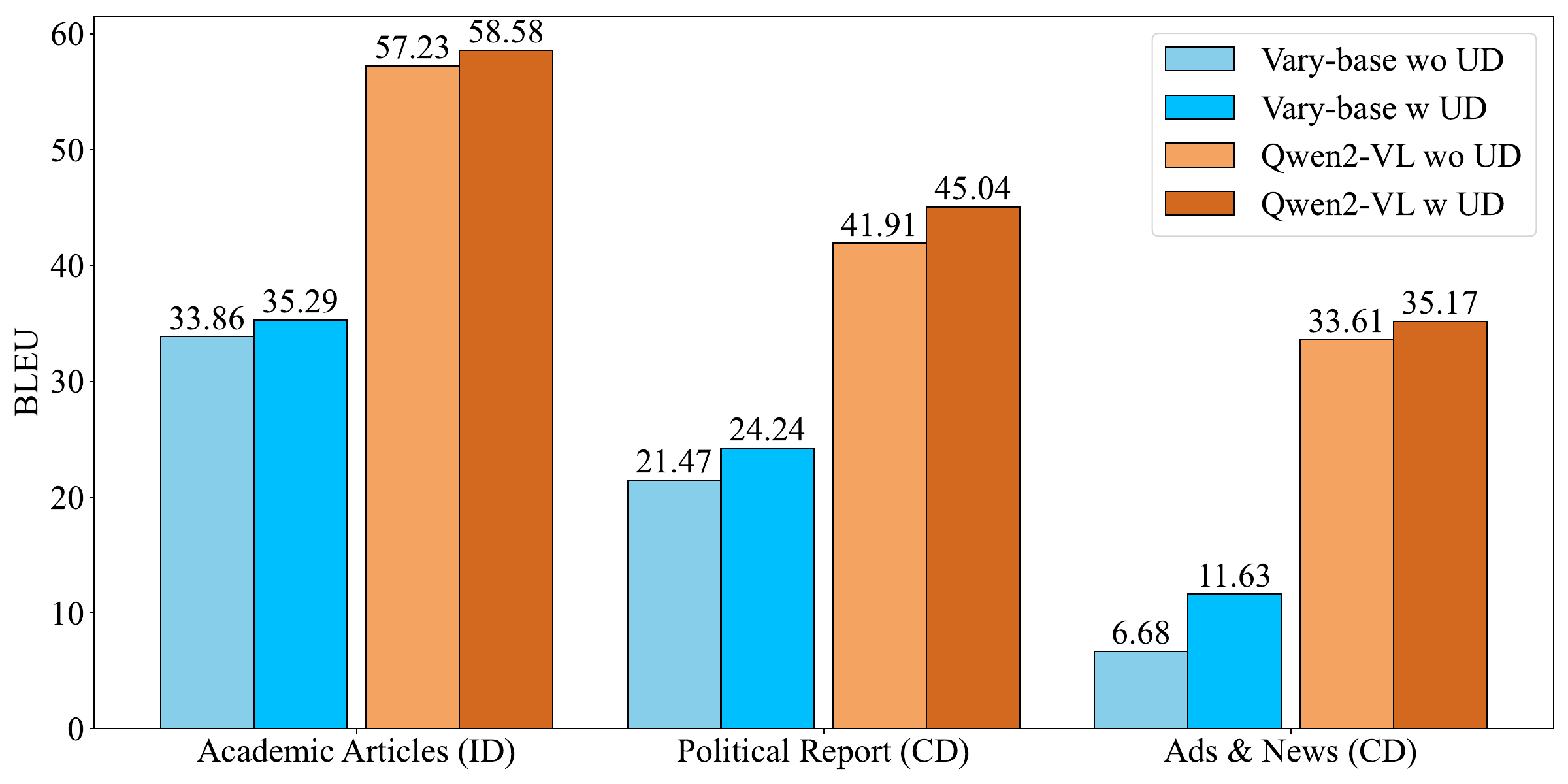}
    \caption{Results of Vary-base and Qwen2-VL through SSR fine-tuning using unsupervised data. \textbf{UD} denotes unsupervised data. Detailed data can be seen in Appendix~\ref{appendix: detailed data}. It is better to zoom in for a clearer view.}
    \label{figure: unsupervised data}
     
\end{figure}

A principal advantage of our method lies in its capacity to harness the MLLM's OCR capability alongside extensive unsupervised data (only document images) to generate synthetic data, thereby augmenting the model's translation performance.
To investigate the effectiveness of incorporating additional unsupervised data, we randomly select 10K document images from the DocVQA training set as the unsupervised data, obtain their OCR results using MLLMs, and translate them into Chinese using \href{https://translate.google.com/}{Google Translate}.
These synthetic data are then integrated with the original training set, and we conduct experiments with Vary-base and Qwen2-VL under the same settings as SSR in the main experiment.
The results are shown in Figure~\ref{figure: unsupervised data}.

As shown in the figure, introducing unsupervised data further enhances the DIMT performance of MLLMs in both in-domain and cross-domain settings compared to the main experiment.
Taking Qwen2-VL as an example, although SSR has already achieved 57.23 BLEU in the academic article domain, our method, which leverages unsupervised data to generate synthetic data, leads to an improvement of 1.35 BLEU in the in-domain test and 3.13 BLEU in the cross-domain test.
This demonstrates the significant potential of our approach for practical applications.

\subsection{Extension to Heterogeneous Source Texts}
Another advantage of our method lies in the extensibility to accommodate heterogeneous source texts.
To validate this capability, we conduct evaluations comparing performance when utilizing ground truth source texts from the DoTA dataset and source texts generated by the \href{https://github.com/madmaze/pytesseract}{OCR tool}.
Experiments are applied to both Vary-base and Qwen2-VL, following the same settings as the main experiment.
The results are listed in Table~\ref{table: extension to GT OCR}.

Table~\ref{table: extension to GT OCR} demonstrates that when the ground truth source text formatting aligns with the MLLM's OCR output format, as observed in Vary-base, SSR achieves performance parity using either ground truth text or self-generated text.
In contrast, significant formatting discrepancies in Qwen2-VL lead to self-generated text consistently outperforming ground truth text in SSR across all evaluated domains.  
Notably, OCR text proves to be a suboptimal variant of ground truth text, with performance degradation attributed to inherent OCR noise artifacts.  
This disparity highlights the importance of format alignment between source texts and the MLLM OCR output for optimal SSR performance.

\section{Conclusion}
In this paper, we propose a novel fine-tuning paradigm, SSR, to enhance MLLMs' DIMT capabilities by leveraging their OCR proficiency, offering three key advantages.
First, monolingual proficiency preserves the MLLM's original monolingual competence by maintaining the source text format.
Second, cross-lingual enhancement enables the MLLM to establish relationships between different modalities, enriching target text generation with additional information.
Finally, our approach can be extended to utilize large-scale unsupervised data to further enhance performance.
Extensive experiments validate the effectiveness of SSR, demonstrating its superiority in strengthening cross-lingual capabilities while preserving monolingual proficiency.

\section*{Limitations}
Although SSR achieves notable results on the DIMT task, its instruction-following ability and user interaction can be further improved.
In the future, we plan to leverage MLLMs' text-grounding capabilities and explore the integration of user prompts to translate text within specific image regions, thereby enhancing translation alignment with user preferences.

\section*{Acknowledgements}
We thank anonymous reviewers for helpful suggestions.
This work is supported by the National Natural Science Foundation of China (No. 62476275 and No. 62476271).

\bibliography{acl_latex}

\clearpage

\appendix
\section*{Appendix}
\section{Setting Details}
\subsection{Dataset Settings}
\label{appendix: dataset settings}
We randomly select 10K samples from the DoTA dataset to form the train set, and use the original valid and test sets.
In the DITrans dataset, the sample sizes for the advertisement, news, and political report subdomains are 485, 610, and 1397, respectively.
Due to the small number of images in the advertisement and news domains and their similar layout structures as scanned document images, we merge these two domains.
We then randomly select 100 images as the test set.
For the political report domain, we also randomly select 100 images as the test set.

\subsection{Main Experiment Settings}
\label{appendix: main experiment settings}
We select four MLLMs with different numbers of parameters: Vary-toy \citep{DBLP:journals/corr/abs-2401-12503}, Vary-base \citep{DBLP:conf/eccv/WeiKCZGYSHZ24}, Textmonkey \citep{DBLP:journals/corr/abs-2403-04473} and Qwen2-VL \citep{DBLP:journals/corr/abs-2409-12191}.
We use the LoRA fine-tuning in our experiments.
The LoRA adapter is added to all the linear layers of the LLM part in the MLLM.
The LoRA rank and alpha are both equal to 16.
We only fine-tune the adapter for 3 epochs with a batch size of 32.
A linear decay learning rate schedule with a learning rate of 1e-4 and a warmup ratio of 0.1 is used.
We use Adam optimizer with $\beta_1 = 0.9$, $\beta_2 = 0.999$, $\epsilon = 1e-8$ for both training stages.
We used two NVIDIA A100 GPUs and spent 16 hours to complete all the training task of SSR in the main experiment.
The greedy search is used for inference.

\subsection{Detailed Prompts}
\label{appendix: prompts}
The OCR instructions used in the main experiment are listed as follows.

\begin{tcolorbox}[colback=lightgray!50!white,colframe=lightgray,title=\textcolor{black}{OCR Instruction for Vary-toy/base}, width=\columnwidth, breakable]
\footnotesize
Convert the document to markdown format.
\end{tcolorbox}

\begin{tcolorbox}[colback=lightgray!50!white,colframe=lightgray,title=\textcolor{black}{OCR Instruction for Textmonkey}, width=\columnwidth, breakable]
\footnotesize
Read all the text in the image.
\end{tcolorbox}

\begin{tcolorbox}[colback=lightgray!50!white,colframe=lightgray,title=\textcolor{black}{OCR Instruction for Qwen2-VL}, width=\columnwidth, breakable]
\footnotesize
Convert the content in the image to Markdown.
\end{tcolorbox}

The instructions of baselines in the main experiment are listed as follows.

\begin{tcolorbox}[colback=lightgray!50!white,colframe=lightgray,title=\textcolor{black}{Instruction for CoT (Direct)}, width=\columnwidth, breakable]
\footnotesize
Convert the content in the image to Markdown \textcolor{gray}{(original OCR instruction of the MLLM)}, then translate into Chinese.
\end{tcolorbox}

\begin{tcolorbox}[colback=lightgray!50!white,colframe=lightgray,title=\textcolor{black}{Prompt Template for CoT (Cascade)}, width=\columnwidth, breakable]
\footnotesize
\texttt{<Round 1>} \newline
\textbf{Instruction}: \newline
Convert the content in the image to Markdown. \textcolor{gray}{(original image caption instruction of the MLLM)} \newline
\newline
\textbf{Response}: \newline
$\bm{X}$ \textcolor{gray}{(self-generated source text)}\newline
\newline
\texttt{<Round 2>} \newline
\textbf{Instruction}: \newline
Translate these text into Chinese. \newline
\newline
\textbf{Response}: \newline
$\bm{Y}$ \textcolor{gray}{(generated target text)}
\end{tcolorbox}

\begin{tcolorbox}[colback=lightgray!50!white,colframe=lightgray,title=\textcolor{black}{Instruction for SFT (DIMT)}, width=\columnwidth, breakable]
\footnotesize
Translate all the text in the image into Chinese and output in Markdown format.
\end{tcolorbox}

The prompt templates used in the monolingual task selection experiment are listed as follows.

\begin{tcolorbox}[colback=lightgray!50!white,colframe=lightgray,title=\textcolor{black}{Prompt Template for Image Caption}, width=\columnwidth, breakable]
\footnotesize
\textbf{Instruction}: \newline
Describe this image \textcolor{gray}{(original image caption instruction of the MLLM)}, then translate into Chinese. \newline
\newline
\textbf{Response}: \newline
$\bm{X}$ \textcolor{gray}{(self-generated image caption text)}\newline
<Translation> \textcolor{gray}{(special token)}\newline
$\bm{Y}$ \textcolor{gray}{(ground truth target text)}
\end{tcolorbox}

\begin{tcolorbox}[colback=lightgray!50!white,colframe=lightgray,title=\textcolor{black}{Prompt Template for VQA}, width=\columnwidth, breakable]
\footnotesize
\textbf{Instruction}: \newline
Convert the content in the image to Markdown \textcolor{gray}{(original OCR instruction of the MLLM)}, then answer the following question: \newline
$\bm{Q}$ \textcolor{gray}{(question from DocVQA, translated into Chinese)} \newline
\newline
\textbf{Response}: \newline
$\bm{X}$ \textcolor{gray}{(self-generated source text)}\newline
<Answer> \textcolor{gray}{(special token)}\newline
$\bm{A}$ \textcolor{gray}{(answer from DocVQA, translated into Chinese)}
\end{tcolorbox}

\begin{table*}[t]
\renewcommand\arraystretch{0.97}
\footnotesize
\centering
\resizebox{\linewidth}{!}{
\begin{tabu}{lcccccccccc}
\toprule
\multicolumn{1}{c|}{\multirow{2}{*}{}} & \multicolumn{3}{c|}{\textbf{Academic Article (ID)}}                    & \multicolumn{3}{c|}{\textbf{Political Report (CD)}}                     & \multicolumn{3}{c|}{\textbf{Ads \& News (CD)}}                          & \textbf{Time}     \\
\multicolumn{1}{c|}{}                  & \textbf{BLEU}  & \textbf{BLEU-PT} & \multicolumn{1}{c|}{\textbf{STEDS}} & \textbf{BLEU}  & \textbf{BLEU-PT} & \multicolumn{1}{c|}{\textbf{STEDS}} & \textbf{BLEU}  & \textbf{BLEU-PT} & \multicolumn{1}{c|}{\textbf{STEDS}} & \textbf{s/page ($\downarrow$)} \\ \midrule

\multicolumn{11}{c}{\textbf{Vary-toy (2.2B)}}                                                                                                                                                                                                                                                     \\ \midrule
\multicolumn{1}{l|}{Base}              & 10.64          & 4.92             & \multicolumn{1}{c|}{66.23}          & 2.07           & 2.10             & \multicolumn{1}{c|}{45.12}          & 0.70           & 0.70             & \multicolumn{1}{c|}{29.60}          & \textbf{43.53}                  \\ \cdashlinelr{1-11}
\multicolumn{1}{l|}{CoT (Direct)}               & 9.17           & 3.87             & \multicolumn{1}{c|}{\textbf{73.45}}          & 2.40           & 2.42             & \multicolumn{1}{c|}{\textbf{59.58}}          & 0.68           & 0.68             & \multicolumn{1}{c|}{\textbf{57.91}}          & 46.88                  \\
\multicolumn{1}{l|}{CoT (Cascade)}               & 3.99           & 1.68             & \multicolumn{1}{c|}{38.13}          & 1.09           & 0.99             & \multicolumn{1}{c|}{36.06}          & 0.23           & 0.27             & \multicolumn{1}{c|}{38.39}          & 62.64                  \\
\multicolumn{1}{l|}{SFT (MT)}          & 1.99           & 1.38             & \multicolumn{1}{c|}{32.14}          & 1.30           & 1.33             & \multicolumn{1}{c|}{41.04}          & 0.47           & 0.47             & \multicolumn{1}{c|}{40.02}          & 185.38                  \\
\multicolumn{1}{l|}{SFT (DIMT)}        & 9.31           & 8.37             & \multicolumn{1}{c|}{62.73}          & 1.49           & 1.47             & \multicolumn{1}{c|}{38.39}          & 0.42           & 0.46             & \multicolumn{1}{c|}{41.06}          &  86.79                 \\
\multicolumn{1}{l|}{SDFT}              & 7.35           & 7.44             & \multicolumn{1}{c|}{57.86}          & 1.54           & 1.56             & \multicolumn{1}{c|}{37.00}          & 0.54           & 0.55             & \multicolumn{1}{c|}{50.79}          & 98.09                  \\
\multicolumn{1}{l|}{SSR}       & \textbf{13.95}          & \textbf{14.21}            & \multicolumn{1}{c|}{65.49}          & \textbf{8.15}           & \textbf{8.22}             & \multicolumn{1}{c|}{49.25}          & \textbf{1.26}           & \textbf{1.34}             & \multicolumn{1}{c|}{42.84}          & 142.29                  \\ \bottomrule
\end{tabu}
}
\caption{Results of different settings of Vary-toy on DoTA and DITrans dataset.}
\label{table: main experiment vary-toy}
\end{table*}

\begin{table*}[t]
\footnotesize
\centering
\begin{tabular}{lccccccccc}
\toprule
\multicolumn{1}{c|}{\multirow{2}{*}{}} & \multicolumn{3}{c|}{\textbf{Academic Articles (ID)}}                   & \multicolumn{3}{c|}{\textbf{Political Report (CD)}}                    & \multicolumn{3}{c}{\textbf{Ads \& News (CD)}}      \\
\multicolumn{1}{c|}{}                  & \textbf{BLEU} & \textbf{BLEU-PT} & \multicolumn{1}{c|}{\textbf{STEDS}} & \textbf{BLEU} & \textbf{BLEU-PT} & \multicolumn{1}{c|}{\textbf{STEDS}} & \textbf{BLEU} & \textbf{BLEU-PT} & \textbf{STEDS} \\ \midrule
\multicolumn{1}{l|}{GPT-4o}            & 29.70          & 31.95            & \multicolumn{1}{c|}{59.45}          & 38.66              & 38.66                 & \multicolumn{1}{c|}{60.54}               & 21.75              & 21.75                 & 59.48               \\
\multicolumn{1}{l|}{Gemini}            & 30.31         & 31.69            & \multicolumn{1}{c|}{63.32}          & 40.11              & 40.11                 & \multicolumn{1}{c|}{69.58}      & 26.83              & 26.83                & 65.31               \\ \midrule
\multicolumn{10}{c}{\textbf{Qwen2-VL (8.3B)}}                                                                                                                                                                                                         \\ \midrule
\rowcolor{lightgray} \multicolumn{1}{l|}{Base}              & 19.56         & 15.38            & \multicolumn{1}{c|}{57.29}          & 26.49         & 26.51            & \multicolumn{1}{c|}{58.10}           & 11.19         & 11.19            & 58.81          \\ \cdashlinelr{1-10}
\multicolumn{1}{l|}{SFT (DIMT)}       & 53.92         & 53.20            & \multicolumn{1}{c|}{87.27}          & 37.96         & 37.93            & \multicolumn{1}{c|}{63.08}          & 23.48         & 23.49            & 69.72          \\
\multicolumn{1}{l|}{SSR}       & \textbf{57.23}         & \textbf{58.88}            & \multicolumn{1}{c|}{\textbf{89.65}}          & \textbf{41.91}         & \textbf{41.80}            & \multicolumn{1}{c|}{\textbf{67.28}}          & \textbf{33.61}         & \textbf{33.59}            & \textbf{71.98}          \\ \bottomrule
\end{tabular}
\caption{Results on comparison with commercial MLLMs. The \textbf{bold numbers} indicate the best performance of all models, including the commercial MLLMs.}
\label{table: commercial mllm}
\end{table*}

\section{Detailed Analysis}

\subsection{Small MLLM Results in the Main Experiment}
\label{appendix: vay-toy}
The results are shown in Table~\ref{table: main experiment vary-toy}. In the Vary-toy experiment, SSR surpasses SFT (DIMT) by 4.64 BLEU on the in-domain test, and also achieves 8.15 BLEU in the political report zero-shot cross-domain test.
These results demonstrate the effectiveness of our method in enhancing both translation quality and generalization in small MLLMs.
Although our method increases inference time, the performance improvement makes this trade-off acceptable.

\begin{table*}[t]
\footnotesize
\centering
\begin{tabular}{lccccccccc}
\toprule
\multicolumn{1}{c|}{\multirow{2}{*}{}} & \multicolumn{3}{c|}{\textbf{Academic Articles (ID)}}                    & \multicolumn{3}{c|}{\textbf{Political Report (CD)}}                     & \multicolumn{3}{c}{\textbf{Ads \& News (CD)}}       \\
\multicolumn{1}{c|}{}                  & \textbf{BLEU}  & \textbf{BLEU-PT} & \multicolumn{1}{c|}{\textbf{STEDS}} & \textbf{BLEU}  & \textbf{BLEU-PT} & \multicolumn{1}{c|}{\textbf{STEDS}} & \textbf{BLEU}  & \textbf{BLEU-PT} & \textbf{STEDS} \\ \midrule
\multicolumn{10}{c}{\textbf{En-Fr} \textbf{Vary-base (8.1B)}}                                                                                                                                                                                                           \\ \midrule
\rowcolor{lightgray} \multicolumn{1}{l|}{Base}          & 18.27          & 11.88            & \multicolumn{1}{c|}{\textbf{81.87}}          & 5.46           & 5.35             & \multicolumn{1}{c|}{\textbf{56.61}}          & 3.85           & 3.85             & \textbf{49.41}          \\ \cdashlinelr{1-10}
\multicolumn{1}{l|}{SFT (DIMT)}        & 30.97          & 29.23            & \multicolumn{1}{c|}{79.16}          & 10.50           & 10.50             & \multicolumn{1}{c|}{48.17}          & 3.42           & 3.41             & 41.03          \\
\multicolumn{1}{l|}{SSR}       & \textbf{44.9}  & \textbf{45.37}            & \multicolumn{1}{c|}{81.71}          & \textbf{35.99} & \textbf{36.68}   & \multicolumn{1}{c|}{56.21}          & \textbf{10.19} & \textbf{10.19}   & 48.79 \\ \midrule
\multicolumn{10}{c}{\textbf{En-Fr} \textbf{Qwen2-VL (8.3B)}}                                                                                                                                                                                                            \\ \midrule
\rowcolor{lightgray} \multicolumn{1}{l|}{Base}          & 30.42          & 27.76            & \multicolumn{1}{c|}{60.22}          & 40.58          & 40.57            & \multicolumn{1}{c|}{60.74}          & 21.16          & 21.16            & 66.56          \\ \cdashlinelr{1-10}
\multicolumn{1}{l|}{SFT (DIMT)}        & 61.19          & 62.81            & \multicolumn{1}{c|}{87.13}          & 41.95          & 41.68            & \multicolumn{1}{c|}{62.21}          & 27.82          & 27.82            & 57.99          \\
\multicolumn{1}{l|}{SSR}       & \textbf{65.25} & \textbf{68.18}   & \multicolumn{1}{c|}{\textbf{89.85}} & \textbf{51.39} & \textbf{51.22}   & \multicolumn{1}{c|}{\textbf{63.34}} & \textbf{38.96} & \textbf{38.97}   & \textbf{66.58} \\ \midrule
\multicolumn{10}{c}{\textbf{En-De} \textbf{Vary-base (8.1B)}}                                                                                                                                                                                                           \\ \midrule
\rowcolor{lightgray} \multicolumn{1}{l|}{Base}          & 18.95          & 12.62            & \multicolumn{1}{c|}{\textbf{82.23}}          & 6.10           & 5.97             & \multicolumn{1}{c|}{56.10}          & 4.47           & 4.47             & \textbf{50.03}          \\ \cdashlinelr{1-10}
\multicolumn{1}{l|}{SFT (DIMT)}        & 29.11          & 27.76            & \multicolumn{1}{c|}{78.35}          & 6.66           & 6.72             & \multicolumn{1}{c|}{48.92}          & 2.73           & 2.87             & 41.84          \\
\multicolumn{1}{l|}{SSR}       & \textbf{38.48} & \textbf{37.66}   & \multicolumn{1}{c|}{79.57}          & \textbf{22.22}          & \textbf{22.76}            & \multicolumn{1}{c|}{\textbf{56.47}}          & \textbf{7.93}  & \textbf{8.41}    & 49.25          \\ \midrule
\multicolumn{10}{c}{\textbf{En-De} \textbf{Qwen2-VL (8.3B)}}                                                                                                                                                                                                            \\ \midrule
\rowcolor{lightgray} \multicolumn{1}{l|}{Base}          & 25.38          & 22.10            & \multicolumn{1}{c|}{60.52}          & 27.23          & 27.28            & \multicolumn{1}{c|}{57.29}          & 19.09          & 19.09            & 64.78          \\ \cdashlinelr{1-10}
\multicolumn{1}{l|}{SFT (DIMT)}        & 56.15          & 55.47            & \multicolumn{1}{c|}{86.40}          & 35.01          & 34.84            & \multicolumn{1}{c|}{61.05}          & 27.18          & 27.18            & 63.52          \\
\multicolumn{1}{l|}{SSR}       & \textbf{58.60} & \textbf{60.32}   & \multicolumn{1}{c|}{\textbf{90.09}} & \textbf{43.43}          & \textbf{43.18}            & \multicolumn{1}{c|}{\textbf{65.61}} & \textbf{27.99}          & \textbf{27.99}            & \textbf{65.31}          \\ \bottomrule
\end{tabular}
\caption{Results on English-French and English-German DIMT test. The \textbf{bold numbers} indicate the best performance of all methods.}
\label{table: other language}
\end{table*}

\subsection{Comparison with Commercial MLLMs}

With the rapid development of MLLMs, some commercial MLLMs \citep{DBLP:journals/corr/abs-2410-21276, DBLP:journals/corr/abs-2403-05530} demonstrate the capability of understanding text-rich document images.
To assess their ability to accomplish the DIMT task, we randomly choose 200 samples from the test set of the DoTA dataset and the original DITrans test sets in the main experiments, then prompt GPT-4o and Gemini with three different prompts to complete the document image machine translation task.
The prompts we used are as follows.

\begin{tcolorbox}[colback=lightgray!50!white,colframe=lightgray,title=\textcolor{black}{Prompts for GPT-4o and Gemini to complete DIMT task}, width=\columnwidth, breakable]
\footnotesize
\texttt{<Prompt 1>} \newline
Output the Chinese translations of this image in markdown format. \newline \newline
\texttt{<Prompt 2>} \newline
Please extract and provide the Chinese translations of the text contained within this image, ensuring that the translations are accurately represented, and format them using markdown for clear presentation. \newline \newline
\texttt{<Prompt 3>} \newline
Please translate the all texts in this image into English and adhere to the following translation standards: \newline
Accuracy: Ensure that the translation fully captures the meaning of all the texts in the image without adding or omitting any information. \newline
Fluency: The translation should read naturally and smoothly, reflecting the conventions of the target language and the translation should follow the reading order of the image. \newline
Format: The translation should be presented in markdown format.
\end{tcolorbox}

We average the metric values of the translation results obtained from different prompts to determine the final results.
As the output format of MLLMs may be unstable, we filter the English parts of the output text and only keep the Chinese parts.

Table~\ref{table: commercial mllm} demonstrates that while GPT-4o and Gemini exhibit inherent capability to execute the DIMT task, surpassing the baseline Qwen2-VL model, they exhibit inferior performance compared to SSR-fine-tuned Qwen2-VL.
This discrepancy stems from commercial MLLMs' lack of training on the DoTA dataset and their divergent output formats relative to the reference standards, resulting in substantially poorer performance on metrics including BLEU and STEDS, compared to Qwen2-VL after fine-tuning.
Notably, Qwen2-VL, after fine-tuning with SSR, maintains superior performance over commercial MLLMs in the political report and ads \& news domains, which are absent from its original training data. 
In contrast, Qwen2-VL fine-tuned with SFT does not exhibit comparable performance.
This comparative analysis substantiates SSR's efficacy in enhancing MLLMs' generalization capabilities for DIMT tasks.

\subsection{Low-resource Scenarios}

\begin{figure}[t]
    \centering
    \includegraphics[width=\columnwidth]{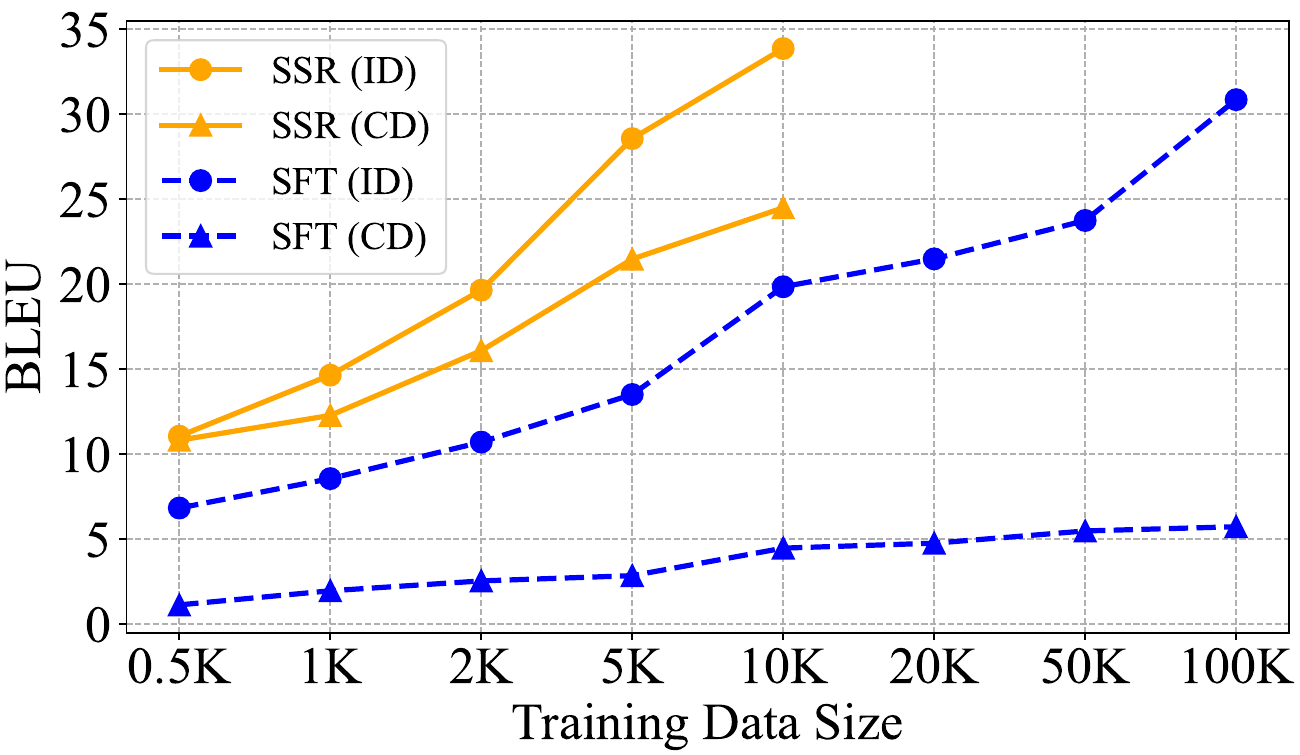}
    \caption{Results of Vary-base through SSR fine-tuning under low-resource scenarios. Detailed data can be seen in Appendix~\ref{appendix: detailed data}.}
    \label{figure: data scale}
     
\end{figure}

\begin{table*}[t]
\centering
\footnotesize
\resizebox{\linewidth}{!}{
\begin{tabu}{l|cccccccc}
\toprule
\multicolumn{1}{c|}{\multirow{2}{*}{}} & \textbf{\begin{tabular}[c]{@{}c@{}}DIMT\\ (Academic Article)\end{tabular}} & \textbf{\begin{tabular}[c]{@{}c@{}}DIMT\\ (Political Report)\end{tabular}} & \textbf{\begin{tabular}[c]{@{}c@{}}DIMT\\ (Ads \& News)\end{tabular}} & \textbf{\begin{tabular}[c]{@{}c@{}}OCR\\ (Document)\end{tabular}} & \textbf{\begin{tabular}[c]{@{}c@{}}OCR\\ (Scene)\end{tabular}} & \textbf{DocVQA} & \textbf{InfoVQA} & \textbf{ChartQA} \\
\multicolumn{1}{c|}{}                  & \textbf{BLEU}                                                              & \textbf{BELU}                                                              & \textbf{BLEU}                                                         & \textbf{CA}                                                       & \textbf{CA}                                                    & \textbf{ANLS}   & \textbf{ANLS}    & \textbf{ANLS}    \\ \midrule
OCR                                    & 57.23                                                                      & 41.91                                                                      & 33.61                                                                 & 85.18                                                             & 82.03                                                          & 92.47           & 60.56            & 61.37            \\
Image Caption                          & 48.20                                                                      & 33.01                                                                      & 20.37                                                                 & 39.80                                                             & 44.83                                                          & 89.98           & 60.62            & 60.36            \\
VQA                                    & 51.70                                                                      & 39.11                                                                      & 28.16                                                                 & 6.53                                                              & 18.03                                                          & 60.12           & 41.43            & 39.63            \\ \bottomrule
\end{tabu}
}
\caption{Detailed data of Figure~\ref{figure: task selection}.}
\label{table: detailed data task selection}
\end{table*}


\begin{table*}[t]
\centering
\footnotesize
\begin{tabu}{lccccccccc}
\toprule
\multicolumn{1}{c|}{\multirow{2}{*}{}} & \multicolumn{3}{c|}{\textbf{Academic Articles (ID)}}                   & \multicolumn{3}{c|}{\textbf{Political Report (CD)}}                    & \multicolumn{3}{c}{\textbf{Ads \& News (CD)}}      \\
\multicolumn{1}{c|}{}                  & \textbf{BLEU} & \textbf{BLEU-PT} & \multicolumn{1}{c|}{\textbf{STEDS}} & \textbf{BLEU} & \textbf{BLEU-PT} & \multicolumn{1}{c|}{\textbf{STEDS}} & \textbf{BLEU} & \textbf{BLEU-PT} & \textbf{STEDS} \\ \midrule
\multicolumn{10}{c}{\textbf{Vary-base (8.1B)}}                                                                                                                                                                                                        \\ \midrule
\multicolumn{1}{l|}{w/o UD}            & 33.86         & 34.50            & \multicolumn{1}{c|}{81.72}          & 21.47         & 22.03            & \multicolumn{1}{c|}{50.92}          & 6.68          & 6.69             & 49.07          \\
\multicolumn{1}{l|}{w UD}              & 35.29         & 37.07            & \multicolumn{1}{c|}{84.61}          & 24.24         & 24.74            & \multicolumn{1}{c|}{52.65}          & 11.63         & 11.56            & 51.54          \\ \midrule
\multicolumn{10}{c}{\textbf{Qwen2-VL (8.3B)}}                                                                                                                                                                                                         \\ \midrule
\multicolumn{1}{l|}{w/o UD}            & 57.23         & 58.88            & \multicolumn{1}{c|}{89.65}          & 41.91         & 41.80            & \multicolumn{1}{c|}{67.28}          & 33.61         & 33.59            & 71.98          \\
\multicolumn{1}{l|}{w UD}              & 58.58         & 60.14            & \multicolumn{1}{c|}{89.94}          & 45.04         & 45.04            & \multicolumn{1}{c|}{63.49}          & 35.17         & 35.17            & 73.12          \\ \bottomrule
\end{tabu}
\caption{Detailed data of Figure~\ref{figure: unsupervised data}. \textbf{UD} denotes unsupervised data.}
\label{table: detailed data unsupervised data}
\end{table*}

To investigate the performance of our method in low-resource scenarios, we fine-tune Vary-base with SSR using different sizes of training data.
The results are presented in Figure~\ref{figure: data scale}.

It can be observed that as the training data size increases, the performance of both methods improves. 
However, SSR consistently outperforms SFT across all data sizes and in both testing scenarios.
With only 10K training samples, SSR surpasses SFT, which utilizes 100K training samples, by 3.01 BLEU on the in-domain test and 18.77 BLEU on the cross-domain test.  
Even with just 500 training samples, SSR still outperforms SFT (100K) by 5.09 BLEU on the cross-domain test, highlighting the exceptional potential of our approach in low-resource scenarios.

\begin{table}[t]
\centering
\footnotesize
\resizebox{\linewidth}{!}{
\begin{tabu}{c|cc|cc}
\toprule
 & \textbf{SFT (ID)} & \textbf{SFT (CD)} & \textbf{SSR (ID)} & \textbf{SSR (CD)} \\ \midrule
\textbf{0.5K}               & 6.82              & 1.12              & 11.05             & 10.81             \\
\textbf{1K}                 & 8.55              & 1.95              & 14.64             & 12.27             \\
\textbf{2K}                 & 10.70             & 2.53              & 19.64             & 16.08             \\
\textbf{5K}                 & 13.50             & 2.84              & 28.56             & 21.47             \\
\textbf{10K}                & 19.84             & 4.46              & 33.86             & 24.49             \\
\textbf{20K}                & 21.48             & 4.75              &                   &                   \\
\textbf{50K}                & 23.74             & 5.47              &                   &                   \\
\textbf{100K}               & 30.85             & 5.72              &                   &                   \\ \bottomrule
\end{tabu}
}
\caption{Detailed data of Figure~\ref{figure: data scale}.}
\label{table: detail data data scale}
\end{table}


\subsection{Evaluation on Other Languages}

To verify our method's effectiveness in other languages, we conduct English-French and English-German DIMT experiments.
We randomly choose 10K samples from the En-Fr and En-De subsets of the DoTA dataset to fine-tune MLLMs.
The rest of the settings remain the same as the main experiment.
The results are shown in Table~\ref{table: other language}.

Taking Qwen2-VL as an example, in the English-French DIMT test, SSR outperforms SFT (DIMT) across all test scenarios, achieving a BLEU score of 65.25 in the in-domain test.  
Similarly, in the English-German DIMT test, SSR surpasses SFT (DIMT) in all test scenarios, reaching a BLEU score of 58.60 in the in-domain test.
These results demonstrate the effectiveness of SSR in enhancing the DIMT capability of MLLMs and improving their generalization in DIMT tasks across different languages.

\section{Detailed Data}
\label{appendix: detailed data}

Table~\ref{table: detailed data task selection} presents the detailed data corresponding to the results of Qwen2-VL through SSR fine-tuning using different monolingual tasks, as shown in Figure~\ref{figure: task selection}.
Table~\ref{table: detailed data unsupervised data} provides the detailed data corresponding to the results of Vary-base and Qwen2-VL through SSR fine-tuning using unsupervised data, as shown in Figure~\ref{figure: unsupervised data}.
Table~\ref{table: detail data data scale} lists the detailed data corresponding to the results of Vary-base through SSR fine-tuning using different training data sizes, as shown in Figure~\ref{figure: data scale}.

\section{Output Samples}
\label{appendix: output samples}

\begin{figure*}[t]
    \centering
    \includegraphics[width=2\columnwidth]{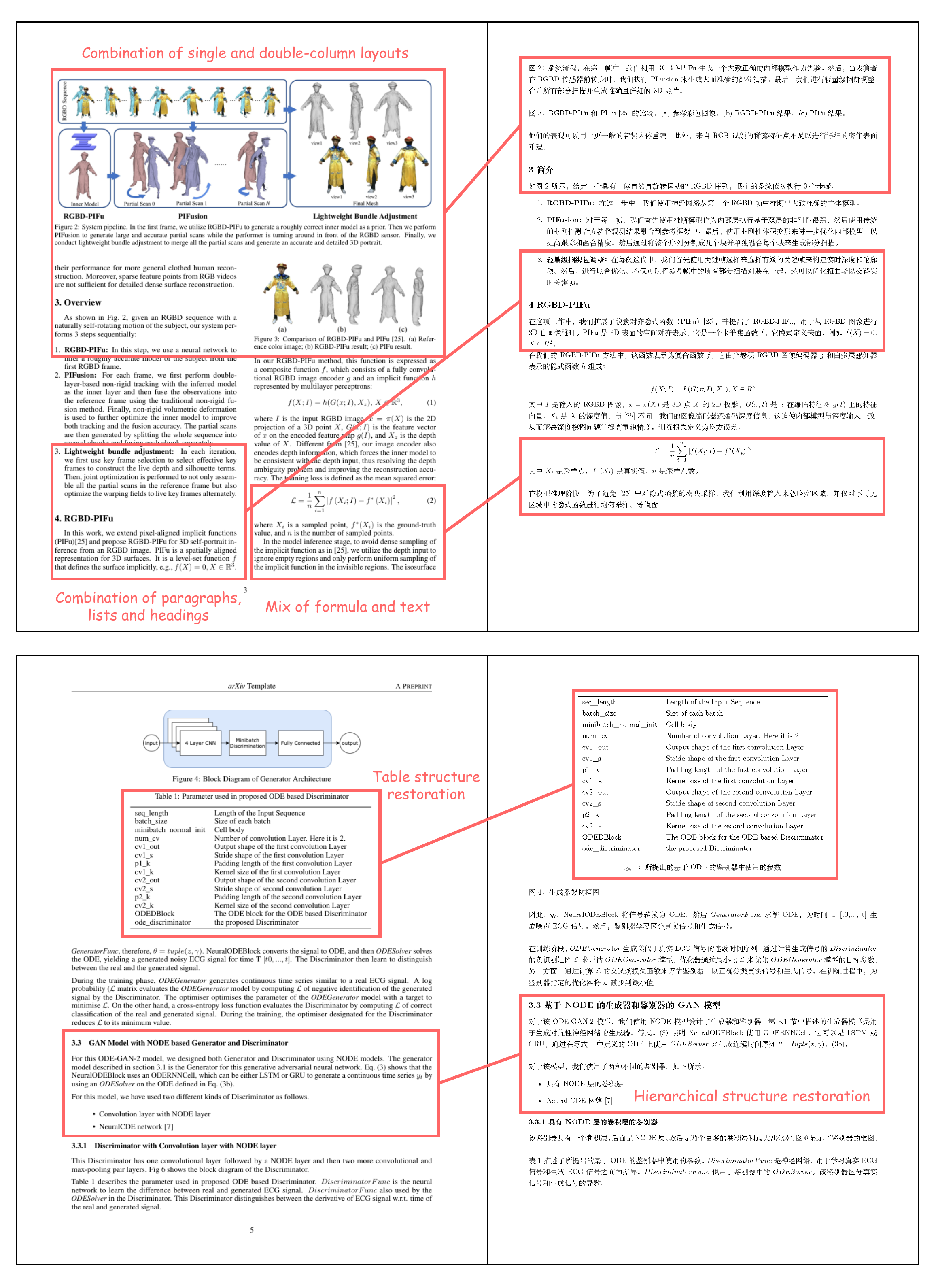}
    \caption{The output samples of Qwen2-VL (after fine-tuning with SSR in the main experiment) on the DoTA test set (Academic Articles). For each image pair, the left is the input document image, and the right is the output translations in markdown format after rendering. It is better to zoom in for a clearer view.}
    \label{figure: case academic}
\end{figure*}

\begin{figure*}[t]
    \centering
    \includegraphics[width=2\columnwidth]{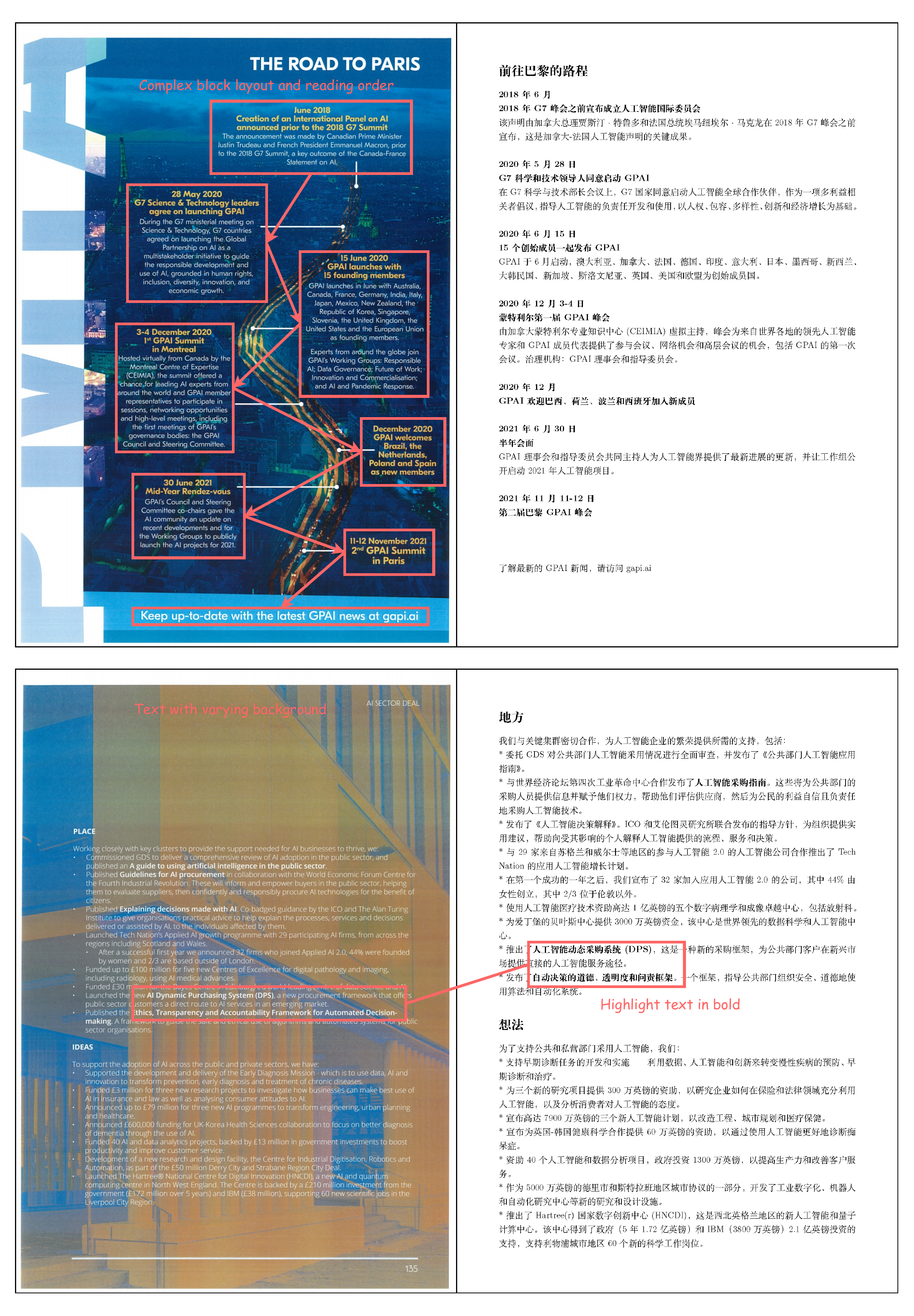}
    \caption{The output samples of Qwen2-VL (after fine-tuning with SSR in the main experiment) on the DITrans test set (Political Report). For each image pair, the left is the input document image, and the right is the output translations in markdown format after rendering. It is better to zoom in for a clearer view.}
    \label{figure: case political}
\end{figure*}

\begin{figure*}[t]
    \centering
    \includegraphics[width=2\columnwidth]{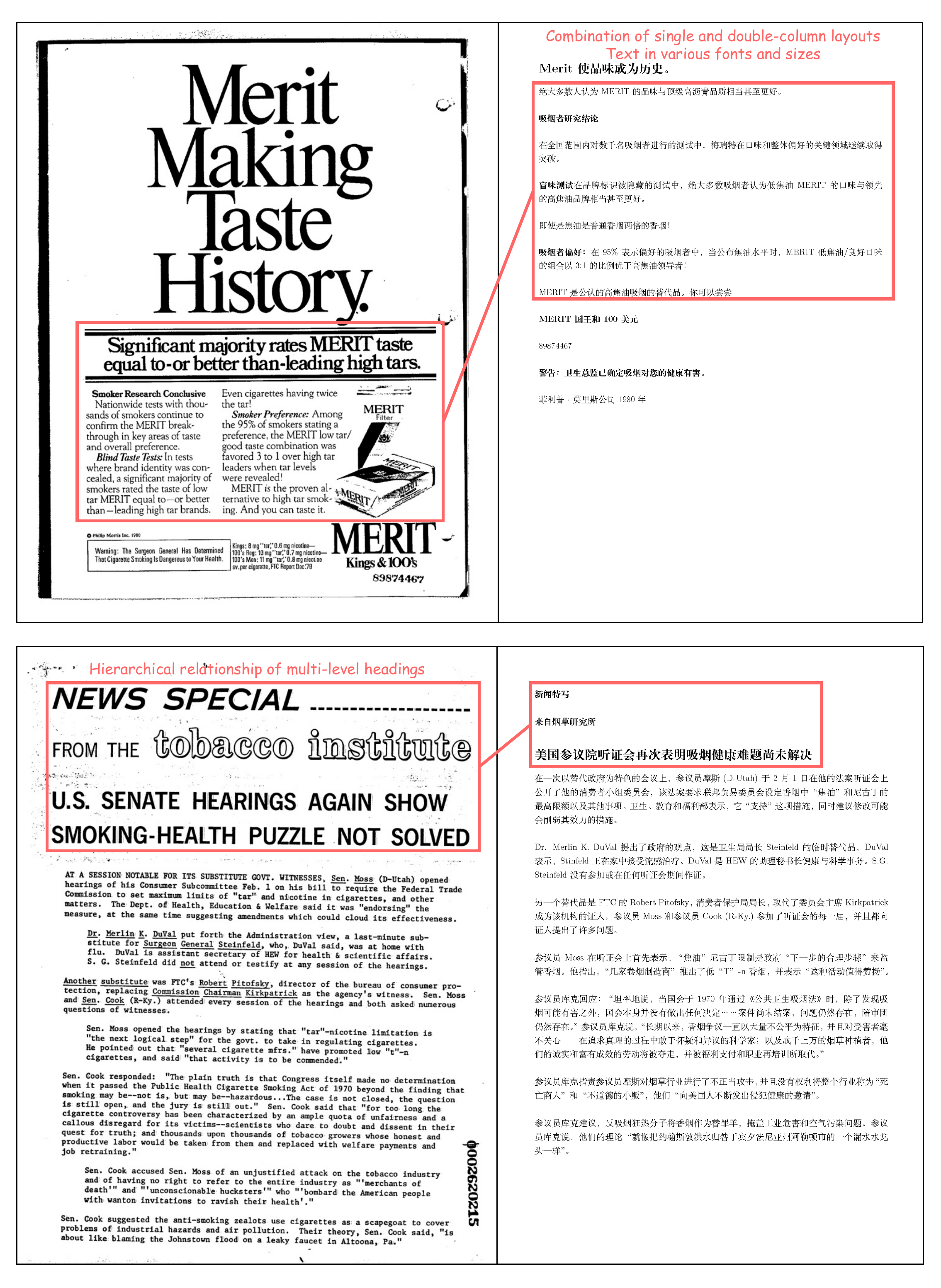}
    \caption{The output samples of Qwen2-VL (after fine-tuning with SSR in the main experiment) on the DITrans test set (Ads \& News). For each image pair, the left is the input document image, and the right is the output translations in markdown format after rendering. It is better to zoom in for a clearer view.}
    \label{figure: case ads}
\end{figure*}

We provide the output samples of Qwen2-VL (after fine-tuning with SSR in the main experiment) on the DIMT test in Figure~\ref{figure: case academic}, Figure~\ref{figure: case political}, and Figure~\ref{figure: case ads}.
It is evident that the MLLM fine-tuned with SSR on the DoTA dataset can understand complex layout relationships and generate translation texts in markdown format following human reading order.
Moreover, it can transfer this capability across domains to political report and ads \& news domains.

\begin{figure*}[t]
    \centering
    \includegraphics[width=2\columnwidth]{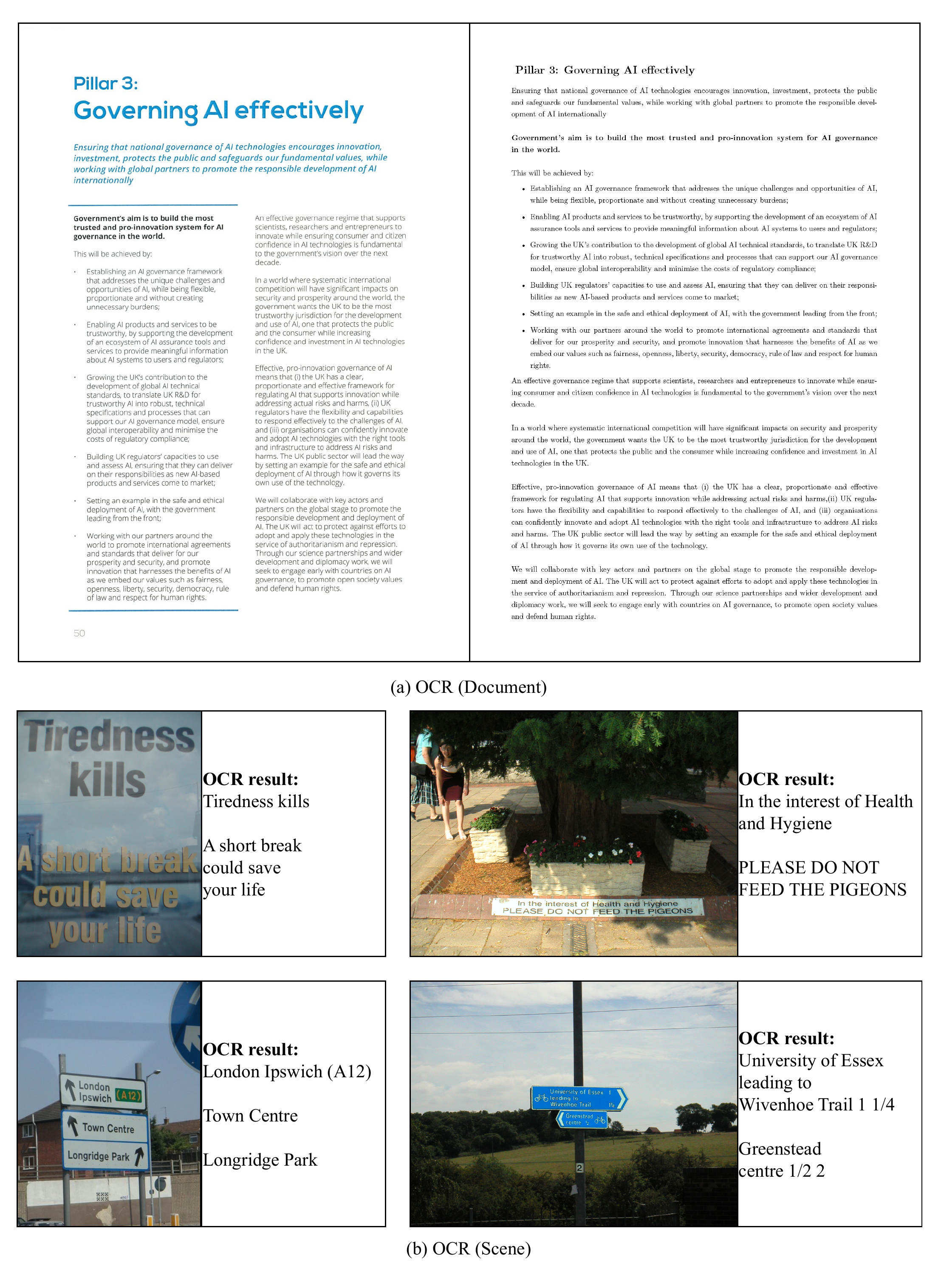}
    \caption{The output samples of Qwen2-VL (after fine-tuning with SSR in the main experiment) on the OCR test. It is better to zoom in for a clearer view.}
    \label{figure: case ocr}
\end{figure*}

\begin{figure*}[t]
    \centering
    \includegraphics[width=2\columnwidth]{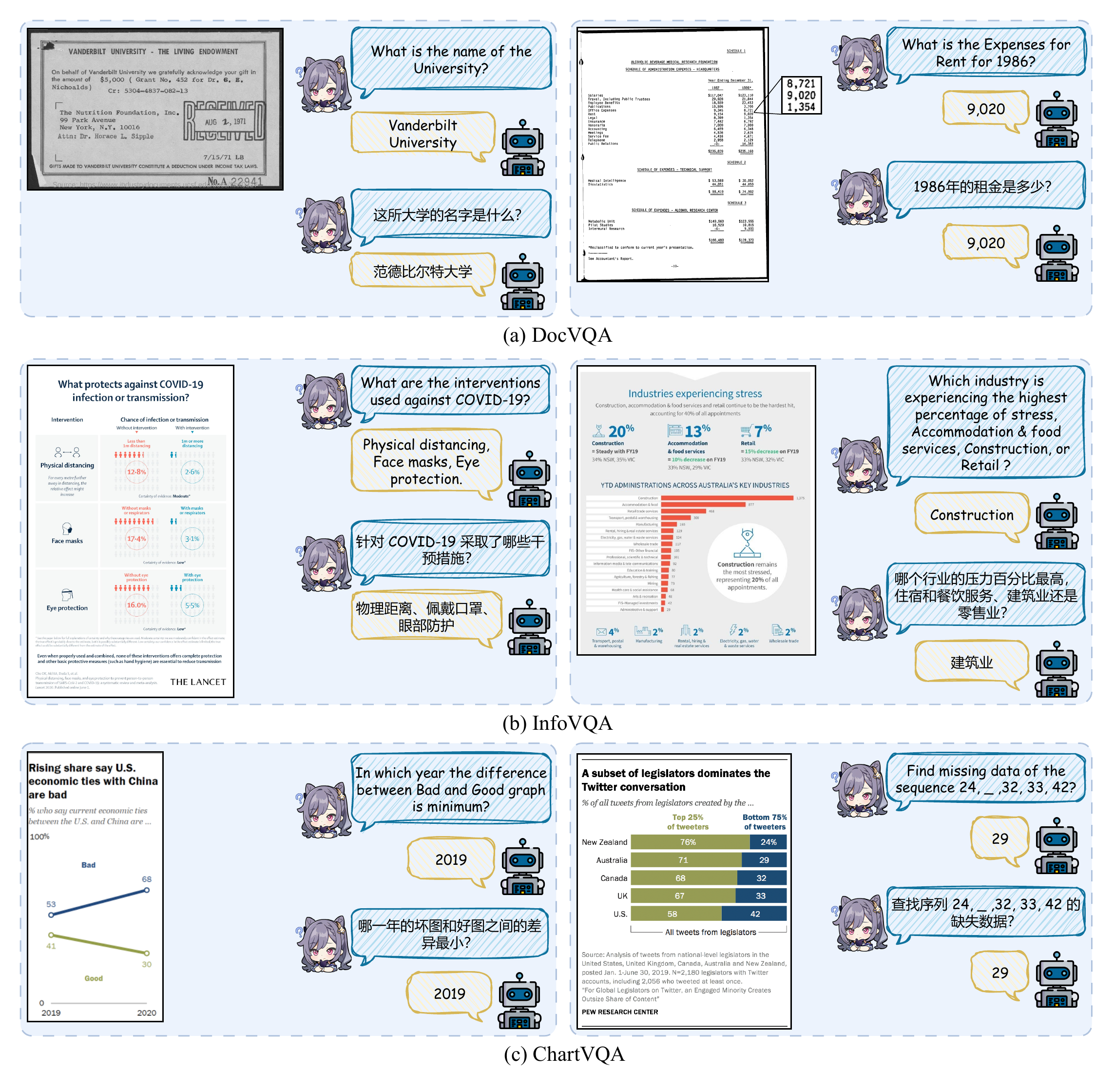}
    \caption{The output samples of Qwen2-VL (after fine-tuning with SSR in the main experiment) on the VQA test. For each document image containing English text, although our model is only trained on the DIMT dataset without utilizing the VQA dataset, it can still respond in the language corresponding to the question. It is better to zoom in for a clearer view.}
    \label{figure: case vqa}
\end{figure*}

Figure~\ref{figure: case ocr} and Figure~\ref{figure: case vqa} show the output samples of Qwen2-VL (after fine-tuning with SSR in the main experiment) on the OCR and VQA test.
As shown in the figure, the MLLM retains strong OCR and VQA capabilities even after being fine-tuned with SSR.
Furthermore, during SSR fine-tuning, the MLLM learns the relationships between English and Chinese, enabling it to generalize cross-lingual VQA capability—allowing it to answer in Chinese when given an English image and a Chinese question.

\end{document}